\pdfoutput=1

\documentclass[twocolumn, switch]{article} 

\usepackage{preprint}

\usepackage{amsmath, amsthm, amssymb, amsfonts}

\usepackage[numbers,square]{natbib}
\usepackage[utf8]{inputenc}	
\usepackage[T1]{fontenc}	
\usepackage{xcolor}		
\usepackage[colorlinks = true,
            linkcolor = purple,
            urlcolor  = blue,
            citecolor = cyan,
            anchorcolor = black]{hyperref}	
\usepackage{booktabs} 		
\usepackage{nicefrac}		
\usepackage{microtype}		
\usepackage{lineno}		
\usepackage{float}			

\usepackage{lipsum}		

\usepackage{newfloat}
\DeclareFloatingEnvironment[name={Supplementary Figure}]{suppfigure}
\usepackage{sidecap}
\sidecaptionvpos{figure}{c}

\usepackage{titlesec}
\titlespacing\section{0pt}{12pt plus 3pt minus 3pt}{1pt plus 1pt minus 1pt}
\titlespacing\subsection{0pt}{10pt plus 3pt minus 3pt}{1pt plus 1pt minus 1pt}
\titlespacing\subsubsection{0pt}{8pt plus 3pt minus 3pt}{1pt plus 1pt minus 1pt}

\usepackage{tikz,xcolor,hyperref}

\definecolor{lime}{HTML}{A6CE39}
\DeclareRobustCommand{\orcidicon}{
	\begin{tikzpicture}
	\draw[lime, fill=lime] (0,0) 
	circle [radius=0.16] 
	node[white] {{\fontfamily{qag}\selectfont \tiny ID}};
	\draw[white, fill=white] (-0.0625,0.095) 
	circle [radius=0.007];
	\end{tikzpicture}
	\hspace{-2mm}
}
\foreach \x in {A, ..., Z}{\expandafter\xdef\csname orcid\x\endcsname{\noexpand\href{https://orcid.org/\csname orcidauthor\x\endcsname}
			{\noexpand\orcidicon}}
}

\title{The Segment Anything Model (SAM) for Remote Sensing Applications: From Zero to One Shot}

\usepackage{authblk}

\author[1\thanks{\tt{lucasosco@unoeste.br}}]{Lucas Prado Osco\orcidA{}}
\author[2]{Qiusheng Wu\orcidE{}}
\author[3]{Eduardo Lopes de Lemos\orcidB{}}
\author[3]{Wesley Nunes Gonçalves\orcidC{}}
\author[4]{Ana Paula Marques Ramos\orcidD{}}
\author[5]{Jonathan Li\orcidF{}}
\author[6]{José Marcato Junior\orcidG{}}

\affil[1]{\scriptsize Faculty of Engineering and Architecture and Urbanism, University of Western São Paulo (UNOESTE), Rod. Raposo Tavares, km 572, Limoeiro, Presidente Prudente 19067-175, SP, Brazil; lucasosco@unoeste.br; pradoosco@gmail.com}
\affil[2]{Department of Geography and Sustainability, University of Tennessee, Knoxville 37996-0925, TN, United States; qwu18@utk.edu}
\affil[3]{Faculty of Computing, Federal University of Mato Grosso do Sul (UFMS), Av. Costa e Silva-Pioneiros, Cidade Universitária, Campo Grande 79070-900, MS, Brazil; lopes.eduardo@ufms.br, wesley.goncalves@ufms.br}
\affil[4]{Departament of Cartography, São Paulo State University (UNESP), Centro Educacional, R. Roberto Simonsen, 305, Presidente Prudente, 19060-900, SP, Brazil; marques.ramos@unesp.br}
\affil[5]{Department of Geography and Environmental Management, University of Waterloo, Waterloo, ON N2L 3G1, Canada; junli@uwaterloo.ca}
\affil[6]{Faculty of Engineering, Architecture and Urbanism and Geography, Federal University of Mato Grosso do Sul (UFMS), Av. Costa e Silva-Pioneiros, Cidade Universitária, Campo Grande 79070-900, MS, Brazil; jose.marcato@ufms.br}

\begin{document}

\twocolumn[ 
  \begin{@twocolumnfalse} 
  
\maketitle

\begin{abstract}
\small Segmentation is an essential step for remote sensing image processing. This study aims to advance the application of the Segment Anything Model (SAM), an innovative image segmentation model by Meta AI, in the field of remote sensing image analysis. SAM is known for its exceptional generalization capabilities and zero-shot learning, making it a promising approach to processing aerial and orbital images from diverse geographical contexts. Our exploration involved testing SAM across multi-scale datasets using various input prompts, such as bounding boxes, individual points, and text descriptors. To enhance the model's performance, we implemented a novel automated technique that combines a text-prompt-derived general example with one-shot training. This adjustment resulted in an improvement in accuracy, underscoring SAM's potential for deployment in remote sensing imagery and reducing the need for manual annotation. Despite the limitations, encountered with lower spatial resolution images, SAM exhibits promising adaptability to remote sensing data analysis. We recommend future research to enhance the model's proficiency through integration with supplementary fine-tuning techniques and other networks. Furthermore, we provide the open-source code of our modifications on online repositories, encouraging further and broader adaptations of SAM to the remote sensing domain.
\end{abstract}
\vspace{0.35cm}

  \end{@twocolumnfalse} 
] 



\section{Introduction}

The field of remote sensing deals with capturing images of the Earth’s surface from airborne or satellite sensors. Analyzing these images allows us to monitor environmental changes, manage disasters, and plan urban areas efficiently \citep{Gmez2016, Song2023, Yuan2020}. A critical part of this analysis is the ability to accurately identify and segment various objects or regions within these images, a process known as image segmentation. Segmentation allows us to isolate specific objects or areas within an image for further study or monitoring \citep{Kotaridis2021}. Traditional segmentation techniques often require extensive human input and intervention for accurate results. However, with the advent of advanced artificial intelligence (AI) and deep learning methods \citep{Bai2022, Aleissaee2023}, the segmentation process has become more automated, albeit still facing challenges, particularly in the effective segmentation of images with minimal human input.

The Segment Anything Model (SAM), developed by Meta AI, is a groundbreaking approach to image segmentation that has demonstrated exceptional generalization capabilities across a diverse range of image datasets, requiring no additional training for unfamiliar objects \citep{kirillov2023segment}. This approach enables it to make accurate predictions with little to no training data. However, its potential can be limited when facing specific domain conditions. To overcome this limitation, SAM can be modified by a re-learning approach \citep{zhang2023personalize}, feeding it with a single example of a new class or object for better results.

Zero-shot learning pertains to a model's capability to accurately process and act upon input data that it hasn’t explicitly encountered during training \citep{alayrac2022flamingo, Sun2021}. This ability is derived from gaining a generalized understanding of the data rather than specific instances. Zero-shot learning systems can recognize objects or understand tasks they have never seen before based on learning underlying concepts or relationships. In contrast, one-shot learning denotes a model's ability to interpret and make accurate inferences from just a single example of a new class \citep{zhang2023personalize}. By feeding SAM with a single example (or 'shot') of this new class, we can potentially enhance its performance, as it has more specific information to work with.

The best-known one-shot methods for SAM are named PerSAM and PerSAM-F, both being training-free personalization approaches \citep{zhang2023personalize}. Given a single image with a reference mask, PerSAM localizes the target concept using a location prior to an initial estimate of where the object of interest is likely to be. The second method is PerSAM-F, a variant of PerSAM that uses one-shot fine-tuning to reduce mask ambiguity. In this case, the entire SAM is frozen (i.e., its parameters are not updated during the fine-tuning process), and two learnable weights are introduced for multi-scale masks. This one-shot fine-tuning variant requires training only two parameters and can be done in as little as ten seconds to enhance performance \citep{zhang2023personalize}. Both are capable of improving SAM, making it a flexible model.

Another important aspect relates to SAM's ability to perform segmentation with minimal input, requiring only a bounding box or a single point as a reference, or even a prompt text as guidance \citep{kirillov2023segment}. This capability has the potential to reduce human labor during the annotation process. Many existing techniques require intensive annotations for each new object of interest, resulting in significant computational overhead and potential delays in time-sensitive applications. SAM, on the other hand, presents an opportunity to alleviate this time-intensive task.

Since SAM's release in April 2023, the geospatial community has shown strong interest in adapting SAM for remote sensing image segmentation. However, a more in-depth investigation is needed. In this context, we present a first-of-its-kind evaluation of SAM, developing both its zero and one-shot learning performance on segmenting remote sensing imagery. We adapted SAM to our data structure, benchmarked it against multiple datasets, and assessed its potential to segment multiscale images. We then evolved SAM's zero-shot characteristic to a one-shot approach and demonstrated that with only one example of a new class, SAM's segmentation performance can be significantly improved.

Our proposal's innovation is within the one-shot technique, which involves using a prompt-text-based segmentation as a training sample (instead of a human-labeled sample), making it an automated process for refining SAM on remote sensing imagery. In this study, we also discuss the implications, limitations, and potential future directions of our findings. Understanding the effectiveness of SAM in this domain is of paramount importance for novel development. In short, with its promise of zero-shot and one-shot learning, SAM has the potential to transform current practices by significantly reducing the time and resources needed for training and annotating data, thereby enabling a quicker, more efficient approach.

\section{Remote Sensing Image Segmentation: A Brief Summary}

The remote sensing field has experienced impressive advancements in recent years, largely driven by improvements in aerial and orbital platform technologies, sensor capabilities, and computational resources \citep{Toth2016, Osco2021}. One of the most critical tasks in remote sensing is image segmentation, which involves partitioning images into multiple segments or regions, each, ideally, corresponding to a specific object or class \citep{Kotaridis2021}. In this section, we focus on providing comprehensive information regarding segmentation processes, deep learning-based methods, and techniques, and explain the overall importance of conducting zero-to-one shot learning.

Traditional image segmentation techniques in remote sensing often rely on pixel-based or object-based approaches. Pixel-based methods, such as clustering and thresholding, involve grouping pixels with similar characteristics, while object-based techniques focus on segmenting images based on properties of larger regions or objects \citep{Hossain2019, Wang2020}. However, these methods can be limited in their ability to handle the complexity, variability, and high spatial resolution of modern remote sensing imagery \citep{Kotaridis2021}.

Segmentation involves various methods designed to separate or group portions of an image based on certain criteria \cite{Zhang2021}. Each method has a unique approach and application. Interactive Segmentation, for example, is a niche within image segmentation that actively incorporates user input to improve the segmentation process, making it more precise and tailored to specific requirements \citep{Li2020, Wu2021}. Different interactive segmentation methods utilize various strategies to include human intelligence in the loop. This makes interactive segmentation particularly useful in tasks where high precision is required, and generic segmentation methods may not suffice.

Super Pixelization is another method that groups pixels in an image into larger units, or "superpixels," based on shared characteristics such as color or texture \citep{Gharibbafghi2018}. This grouping can simplify the image data while preserving the essential structure of the objects. Object Proposal Generation goes a step further by suggesting potential object bounding boxes or regions within an image \citep{Hossain2019, Su2019}. These proposals serve as a guide for a more advanced model to identify and classify the actual objects' pixels. Foreground Segmentation, also known as background subtraction, is a technique primarily used to separate the main subjects or objects of interest (the foreground) from the backdrop (the background) in an image \citep{Zheng_2020_CVPR, Ma2022}. 

Semantic Segmentation is a more comprehensive approach where every pixel in an image is assigned to a specific class, effectively grouping regions of the image based on semantic interest \citep{Zhang2020b, Adam2023}. Instance Segmentation identifies each pixel recognizes distinct objects of the same class and recognizes the individual objects as separate entities or instances \citep{Gao2021, Qurratulain2023}. Panoptic Segmentation merges the concepts of semantic and instance segmentation, assigning every pixel in the image a class label and a unique instance identifier \citep{Hua2021, deCarvalho2022}. This method aims to give a complete understanding of the image by identifying and classifying every detail.

All these methods have been intensively studied, but one that surged in recent years, with the advancements of Visual Foundation Models (VFM) and Large Multimodal Models (LMM), is known as "Promptable Segmentation," an approach that aims to create a versatile model capable of adapting to a variety of segmentation tasks \citep{mialon2023augmented, zhang2023visionlanguage}. This is achieved through "prompt engineering," where prompts are carefully designed to guide the model toward generating the desired output \citep{Lobry2020, Sun2021}. This concept is a departure from traditional multi-task systems where a single model is trained to perform a fixed set of tasks. The unique feature of a promptable segmentation model is its ability to take on new tasks at the time of inference, serving as a component in a larger system \citep{Sun2021, mialon2023augmented}. For instance, to perform instance segmentation, a promptable segmentation model could be combined with an existing object detector.

Object detection is a crucial task in computer vision, focusing on identifying and locating objects within images. This task is foundational for various applications such as surveillance, autonomous vehicles, and many others. In the realm of object detection and image segmentation, different techniques have been employed. Traditional methods often focus on detecting objects that the model has been specifically trained on, known as closed-set detection. However, real-world applications demand more flexibility and the ability to detect and classify objects not seen during training, known as open-set detection.

One state-of-the-art open-set object detector that stands out is Grounding DINO (GroundDINO), an enhanced transformer-based object detector capable of identifying a broader range of objects based on various human inputs \citep{liu2023grounding}. This system is an enhancement of the Transformer-based object detector called DINO \citep{zhang2022dino}, enriched with grounded pre-training to be able to identify a broader range of objects based on human inputs, such as category names or referring expressions. An open-set detector is meant to identify and classify objects that weren't part of the model's training data, as opposed to a closed-set detector that can only recognize objects it has been specifically trained on. The information from Grounding DINO can potentially be used to guide the segmentation process, providing class labels or object boundaries that the segmentation model could use.

Most NLMs incorporate deep-learning-based networks and, with the rise of these methods, more advanced segmentation techniques have been developed for remote sensing applications. Convolutional Neural Networks (CNNs), which emerged as a popular choice due to their ability to capture local and hierarchical patterns in images \citep{Martins2021, Bressan2022}, have widely been used as the backbone for these tasks. CNNs consist of multiple convolutional layers that apply filters to learn increasingly complex features, making them well-suited for segmenting objects in many remote sensing images \citep{Yuan2021, Bai2022}. However, they are computationally intensive and may require substantial training data.

Generative Adversarial Networks (GANs) have also shown potential in the field of image processing. GANs consist of a generator and a discriminator network, where the generator tries to create synthetic data to fool the discriminator, and the discriminator aims to distinguish between real and synthetic data \citep{Jozdani2022}. For image segmentation, GANs can be used to generate realistic images and their corresponding segmentations, which can supplement the training data and improve the robustness of the segmentation models \citep{BenjdiraRS2019}.

Vision Transformer (ViT), on the other hand, is a recent development in deep learning that has shown promise in image segmentation tasks. Unlike CNNs, which rely on convolutional operations, ViT employs self-attention mechanisms that allow it to model long-range dependencies and global context within images \citep{li2023uniformer, li2023transformerbased}. This approach has demonstrated competitive performance in various computer vision tasks, including remote sensing image segmentation \citep{Aleissaee2023}, and it is currently outperforming CNNs in remote sensing data \citep{Gonalves2023}.

Another capability of deep learning that can enhance the segmentation process is transfer learning. With it, a model pre-trained on a large dataset is adapted for a different but related task \citep{Tong2020}. For instance, a CNN or ViTr trained on a large-scale image recognition dataset like ImageNet can be fine-tuned for the task of remote sensing image segmentation \citep{Osco2020, Osco2021segmentation}. The advantage of transfer learning is that it can leverage the knowledge gained from the initial task to improve performance on the new task, especially when the amount of labeled data for the new task is limited.

One of the main challenges in applying deep learning techniques to remote sensing image segmentation is the need for large volumes of labeled ground-truth data \citep{Chi2016}. Acquiring and annotating this data can be time-consuming and labor-intensive, requiring expert knowledge and resources that may not be readily available. Furthermore, the variability and complexity of remote sensing imagery can make the labeling process even more difficult \citep{Amani2020}. As such, it becomes imperative to develop robust, efficient, and accessible solutions that can aid in the processing and analysis of such data. A model that can perform segmentation with zero domain-specific information may offer an important advantage for this process.

In this sense, the Segment Anything Model (SAM) has emerged as a potential tool for assisting in the segmentation process of remote sensing images. SAM design enables it to generalize to new image distributions and tasks effectively and already resulted in numerous applications \citep{kirillov2023segment}. By using minimal human input, such as bounding boxes, reference points, or simply text-based prompts, SAM can perform segmentation tasks without requiring extensive ground-truth data. This capability can reduce the labor-intensive process of manual annotation and be incorporated into the image processing pipeline, potentially accelerating its workflow.

SAM has been trained on an enormous dataset, of 11 million images and 1.1 billion masks, and it boasts impressive zero-shot performance on already a variety of segmentation tasks \citep{kirillov2023segment}. Foundation models such as this, which have shown promising advancements in NLP and, more recently, in computer vision, can carry out zero-shot learning. This means they can learn from new datasets and perform new tasks often by utilizing 'prompting' techniques, even with little to no previous exposure to these tasks. In the field of NLP, "foundation models" refer to large-scale models that are pre-trained on a vast amount of data and are then fine-tuned for specific tasks. These models serve as the "foundation" for various applications \citep{mai2023opportunities, mialon2023augmented, wu2023visual}. 

SAM's ability to generalize across a wide range of objects and images makes it particularly appealing for remote sensing applications. That it can be retrained with a single example of each new class at the time of prediction \citep{zhang2023personalize}, demonstrates the models' high flexibility and adaptability. The implementation of a one-shot approach may assist in designing models that learn useful information from a small number of examples – in contrast to traditional models which usually require large amounts of data to generalize effectively. This could potentially revolutionize how we process remote-sensing imagery. As such, by investigating SAM's innovative technology, we may be able to provide more interactive and adaptable remote sensing systems.

\section{Materials and Methods}

\begin{figure*}[ht!]
\centering
\includegraphics[width=\textwidth]{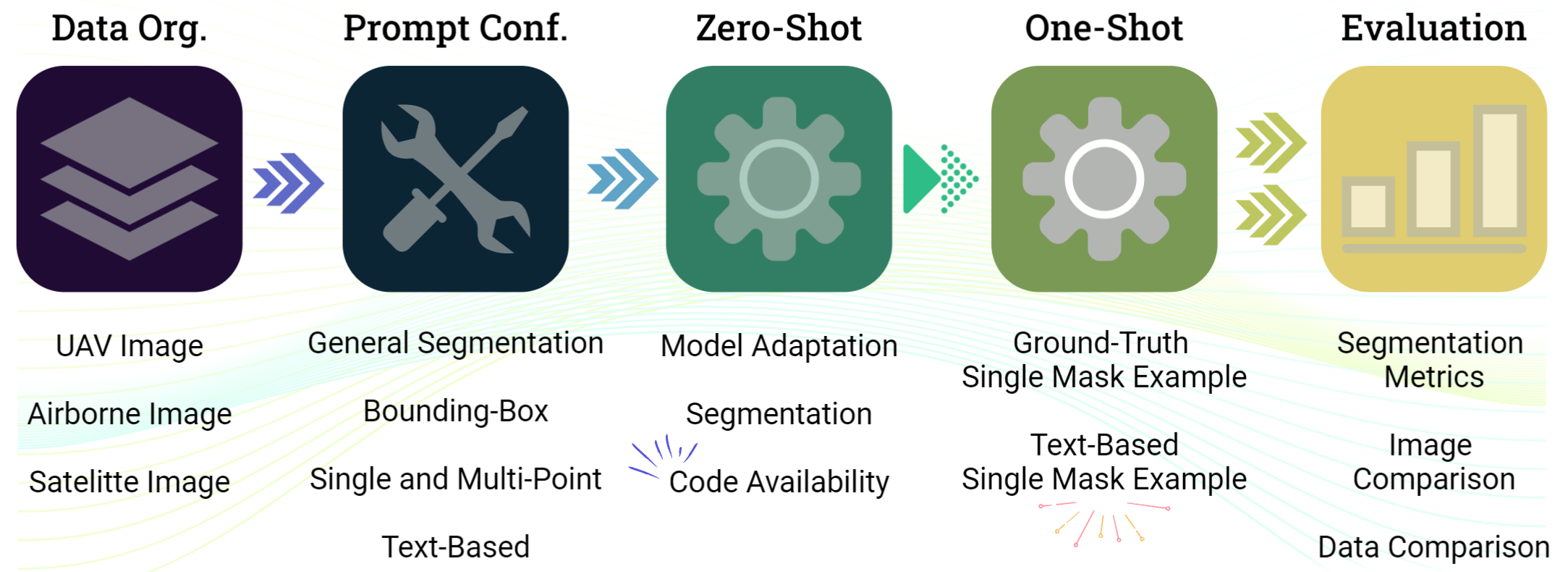}
\caption{\small \centering Schematic representation of the step-by-step process undertaken in this study to evaluate the efficacy of SAM's approach in remote sensing image processing tasks. \label{fig_method}}
\end{figure*}

In this section, we describe how we evaluated the performance of the Segment Anything Model (SAM), for both zero and one-shot approach, in the context of remote sensing imagery. The method implemented in this study is summarized in Figure~\ref{fig_method}. The data for this study consisted of multiple aerial and satellite datasets. These datasets were selected to ensure diverse scenarios and a large range of objects and landscapes. This helped in assessing the robustness of SAM and its adaptability to different situations and geographical regions.

The study particularly investigated SAM's segmentation capacity under different prompting conditions. First, we used the general segmentation approach, in which SAM was tasked to segment objects and landscapes without any guiding prompts. This provided a baseline for SAM's inherent segmentation capabilities with zero-shot. For this, we only evaluated its visual quality, since it segments every possible object in the image, instead of just the ones with ground-truth labels. It also is not guided by any means, thus resulting in the segmentation of unknown classes, serving as just a traditional segmentation filter. 

In the second scenario, bounding boxes were provided. These rectangular boxes, highlighting specific areas within the images, were used to restrict SAM's segmentation per object and see its proficiency in recognizing and segmenting them. Next, we conducted segmentation using points as prompts. In this setup, a series of specific points within the images were provided to guide SAM’s processing. It allowed us to test the precision potential of SAM. Finally, we experimented with the segmentation process using only textual descriptions as prompts. This was conducted with an implementation of SAM alongside GroundingDINO's method \citep{liu2023grounding}. This permitted an evaluation of these models' capabilities to understand, interpret, and transform textual inputs into precise segmentation outputs.

To measure SAM's adaptability and potential to deal with remote sensing imagery, we then devised a one-shot implementation. For each of the datasets, we presented an example of the target class to SAM. For that, we adapted the model with a novel combination of the text-prompt approach and the one-shot learning method. Specifically, we selected the best possible example (highest logits) of the target object, using textual prompts to define the object for mask generation. This example was then presented to SAM as the sole representative of the class, effectively guiding its learning process. The rationale behind this combined approach was to leverage the context provided by the text prompts and the efficacy of the one-shot learning method to the adaptability of SAM to an automated enhancement process.

\subsection{Description of the Datasets}

\begin{figure}[ht!]
\centering
\includegraphics[width=\columnwidth]{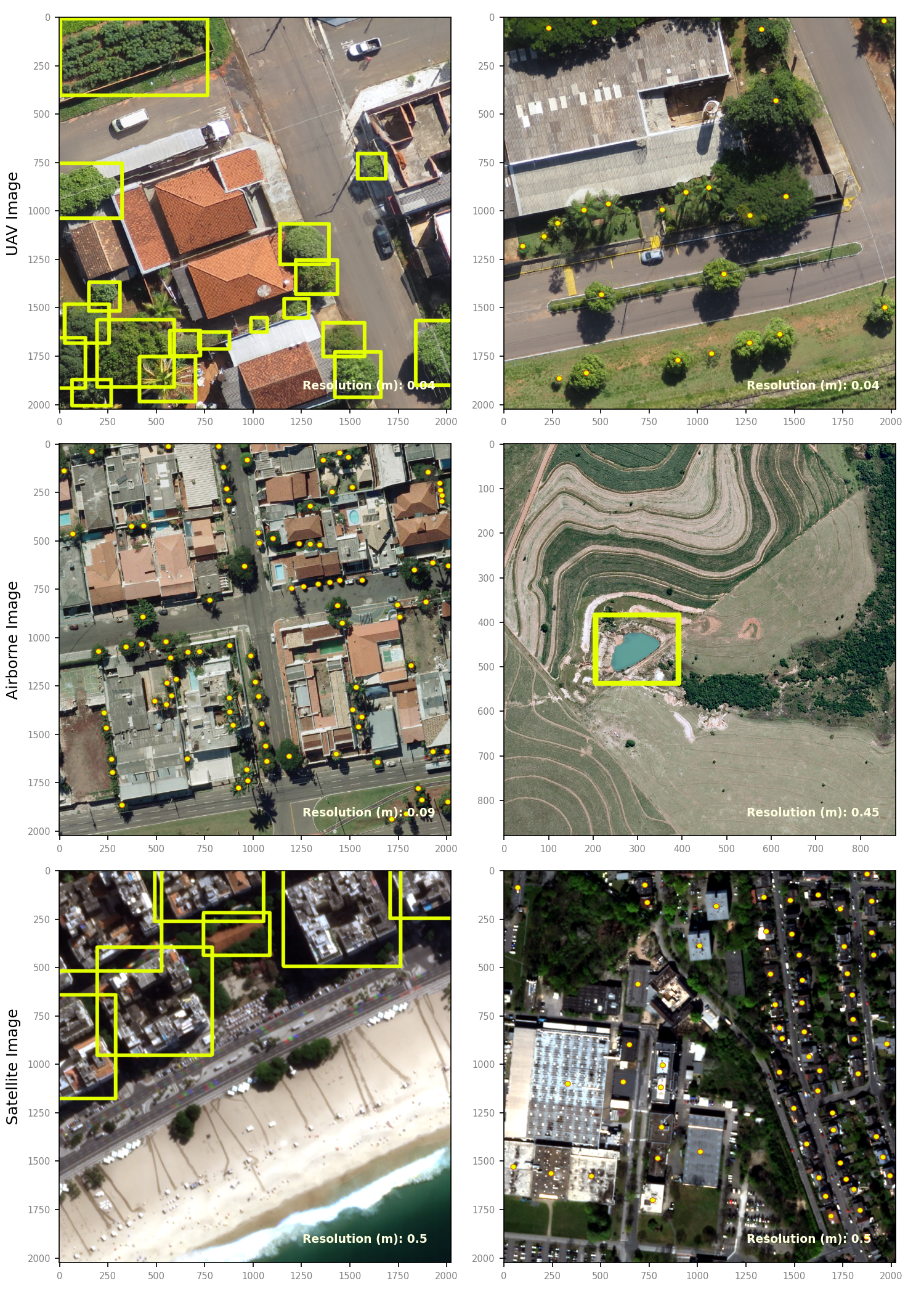}
\caption{\small \centering  Collection of image samples utilized in our research. The top row features UAV-based imagery with bounding boxes and point labels, serving as prompts for SAM. The middle row displays airborne-captured data representing larger regions, with both points and a rectangular box provided as model inputs. The bottom row reveals satellite imagery, again with bounding boxes and points as prompt inputs, offering a trade-off between lower spatial resolution and wider area coverage. \label{fig_data}}
\end{figure}

We begin by separating our dataset into three categories related to the platform used for capturing the images: 1. Unmanned Aerial Vehicle (UAV); 2. Airborne, and; 3. Satellite. Each of these categories provides unique advantages and challenges in terms of spatial resolution and coverage. In our study, we aim to evaluate the performance of SAM across these sources to understand its applicability and limitations in diverse contexts. Their characteristics are summarized in Table~\ref{tab_data}. We also provided illustrative examples from these datasets in Figure~\ref{fig_data} as in bounding boxes and point prompts.

\begin{table*}[ht!]
\centering
\caption{\small \centering Overview of the distinct attributes and specifications of the datasets employed in this study.}
\label{tab_data}
\resizebox{\textwidth}{!}{%
\begin{tabular}{cccccccccc}
\hline
\textbf{\#} &\textbf{Platform} & \textbf{Resolution} & \textbf{Area} & \textbf{Target} & \textbf{General} & \textbf{Box} & \textbf{Point} & \textbf{Text} & \textbf{Reference} \\ \hline
00 & UAV & 0.04 m & 70 ha & Tree & Yes & Yes & Centroid & Tree &  \\
01 & UAV & 0.04 m & 70 ha & House & Yes & Yes & Centroid & House &  \\
02 & UAV & 0.01 m & 4 ha & Plantation Crop & Yes & No & Multiple & Plantation & \cite{Osco2021_cnn} \\
03 & UAV & 0.04 m & 40 ha & Plantation Crop & Yes & No & Multiple & Plantation &  \\
04 & UAV & 0.09 m & 90 ha & Building & Yes & Yes & Centroid & Building & \cite{Gao2021} \\
05 & UAV & 0.09 m & 90 ha & Car & Yes & Yes & Centroid & Car &  \\
06 & Airborne & 0.20 m & 120 ha & Tree & Yes & Yes & Centroid & Tree &  \\
07 & Airborne & 0.20 m & 120 ha & Vehicle & Yes & Yes & Centroid & Vehicle &  \\
08 & Airborne & 0.45 m & 190 ha & Lake & Yes & Yes & Centroid & Lake &  \\
09 & Satelitte & 0.30 m & -- & Building; Road; Water; Barren; Forest; Farm & Yes & Yes & Multiple & Building; Road; Water; Barren; Forest; Farm & LoveDA \cite{wang2022loveda} \\
10 & Satelitte & 0.50 m & 480 ha & Building; Street; Water; Vehicle; Tree & Yes & Yes & Yes & Building; Street; Water; Vehicle; Tree & SkySat ESA \cite{esa2023} \\ \hline
\end{tabular}%
}
\end{table*}

The UAV category comprises data that have the advantage of very-high spatial resolution, returning images and targets with fine details. This makes them particularly suitable for local-scale studies and applications that require high-precision data. However, the coverage area of UAV datasets is limited compared to other data sources. The images comprised particularly single-class objects per dataset, so they were tackled in binary form. In the case of linear objects, specifically continued plantation crops cover, we used multi-points spread within its extremes, to ensure that the model was capable of understating it better. For more condensed targets such as houses and trees, we used the centered position of the object as a point prompt.

The second category is Airborne data, which includes data collected by manned aircraft. These datasets typically offer a good compromise between spatial resolution and coverage area. We processed these datasets with the same approach as with the UAV images since they also consisted of binary problems. The total quantifiable size of these datasets surpasses 90 Gigabytes and comprises more than 10,000 images and image patches. Part of the dataset, specifically the aerial one (UAV and Airborne), is currently being made public in the following link for others to use: \href{https://sites.google.com/view/geomatics-and-computer-vision/home/datasets}{Geomatics and Computer Vision/Datasets}. These datasets cover different area sizes and their corresponding ground-truth masks were generated and validated by specialists in the field.

The third category consists of Satellite data, which provides the widest coverage and is focused on multi-class problems. The spatial resolution of satellite data is generally lower than that of UAV and Airborne data. Furthermore, the quality of the images is more affected by atmospheric conditions, with differing illumination conditions, thus providing additional challenges for the model. These datasets consist of publicly available images from the LoveDA dataset \citep{wang2022loveda} and from the SkySat ESA archive \citep{esa2023} and present a multi-class segmentation problem. To facilitate's SAM evaluation, specifically with the guided prompts (bounding box, point, and text), we conducted a one-against-all approach, in which we separated the classes into individual classifications ("specified class" versus "background").

\subsection{Protocol for Promptable Image Segmentation}

In this section, we explain how we adapted SAM to the remote sensing domain and how we conducted the prompable image segmentation with it. All of the implemented code, specifically designed for this paper, is made publicly available in an under-construction educational repository \citep{AIRemoteSensing2023}. Also, as part of our work, we are focusing on developing the "segment-geospatial" package \citep{segment-geospatial2023}, which implements features that will simplify the process of using SAM models for geospatial data analysis. This is a work in progress, but it is publicly available and offers a suite of tools for performing general segmentation on remote-sensing images using SAM. The goal is to enable users to engage with this technology with a minimum of coding effort.

Our geospatial analysis was conducted with the assistance of a custom tool, namely "SamGeo", which is a component of the original module. SAM possesses different models to be used, namely: ViT-H, ViT-L, and ViT-B \citep{kirillov2023segment}. These models have different computational requirements and are distinct in their underlying architecture. In this study, we used the ViT-H SAM model, which is the most advanced and complex model currently available, bringing most of the SAM capabilities to our tests.

To perform the general prompting, we used the generate method of the SamGeo instance. This operation is simple enough since it segments the entire image and stores it as an image mask file, which contained the segmentation masks. Each mask delineates the foreground of the image, with each distinct mask allocated a unique value. This allowed us to segment different geospatial features. The result is a non-classified segmented image that can also be converted into a vector shape. As mentioned, we only evaluated this approach visually, since it was not possible to appropriately assign the segmented regions outside of our reference class.

For the bounding box prompt, we used the SamGeo instance in conjunction with the objects' shapefile. Bounding boxes are extracted from any multipart polygon geometry returning a, which returned a list of geometric boundaries for our image data based on its coordinates. To efficiently process these boundaries, we initialized the predictor instance. In this process, the image was segmented and passed through the predictor along with a designated model checkpoint. Once established, the predictor processed each clip box, creating the masks for the segmented regions. This process enabled each bounding box's contents to be individually examined as instance segmentation masks. These binary masks were then merged and saved as a single mosaic raster to create a comprehensive visual representation of the segmented regions. Although not focused on remote sensing data, the official implementation is named Grounded-SAM \citep{groundedsam2023}.

The single-point feature prompt was implemented similarly to the bounding-box method. For that, we first defined functions to convert the geodata frame into a list of coordinates [x, y] instead of the previous [x1, y1, x2, y2] ones. We utilized SamGeo again for model prediction but with the distinction of setting its automatic parameter to 'False' and applying the predictor to individual coordinates instead of the bounding boxes. This approach was conducted by iterating through each point, predicting its features in instances, and saving the resulting mask into a unique file per point (also resulting in instance segmentation masks). After the mask files were generated, we proceeded to merge these masks into a single mosaic raster file, giving us a complete representation of all the segmented regions from the single-point feature prompt.

The text-based prompt differentiates from the previous approach since it required additional steps to be implemented. This method combines GroundingDINO's \citep{liu2023grounding} capabilities for zero-shot visual grounding with SAM's object segmentation functionality for retrieving the pre-trained models. For instance, once Grounding DINO has detected and classified an object, SAM is used to isolate that object from the rest. As a result, we've been able to identify and segment objects within our images based on a specified textual prompt. This procedure opens up a new paradigm in geospatial analysis, harnessing the power of state-of-the-art models to extract image features based only on natural language input. 

Since remote sensing imagery often contained multiple instances of the same object (e.g., several 'houses', 'cars', 'trees', etc.), we've added a looping procedure. The loop identifies the object with the highest probability in the image (i.e. logits), creates a mask for it, removes it from the image, and then restarts the process to identify the next highest probable object. This process continues until the model reaches a defined minimum threshold for both detection, based on a box threshold, and text prompt association, also based on an specific threshold. The precise balancing of these thresholds (randing from 0 to 1) is crucial, with implications for the accuracy of the model, so we manually set them for each dataset based on trial and error tentatively:

\begin{itemize}
    \item Box Threshold: Utilized for object detection in images. A higher value augments model selectivity, isolating only those instances the model identifies with high confidence. A lower value, conversely, expands model tolerance, enhancing overall detections but possibly including less certain ones.
    \item Text Threshold: Utilized for associating detected objects with provided text prompts. An elevated value mandates a robust association between the object and text, ensuring precision but potentially limiting associations. A diminished value permits broader associations, potentially boosting the number of associations but potentially compromising precision.
\end{itemize}

These thresholds are critical for ensuring the balance between precision and recall based on specific data and user requirements. The optimal values may diverge depending on the nature and quality of the images and the specificity of text prompts, warranting user experimentation for optimal performance. The segmented individual images and their corresponding boxes are subsequently generated, while the resulting segmentation mask is saved and mosaicked.

\subsection{One-Shot Text-Based Approach}

\begin{figure*}[ht!]
\centering
\includegraphics[width=\textwidth]{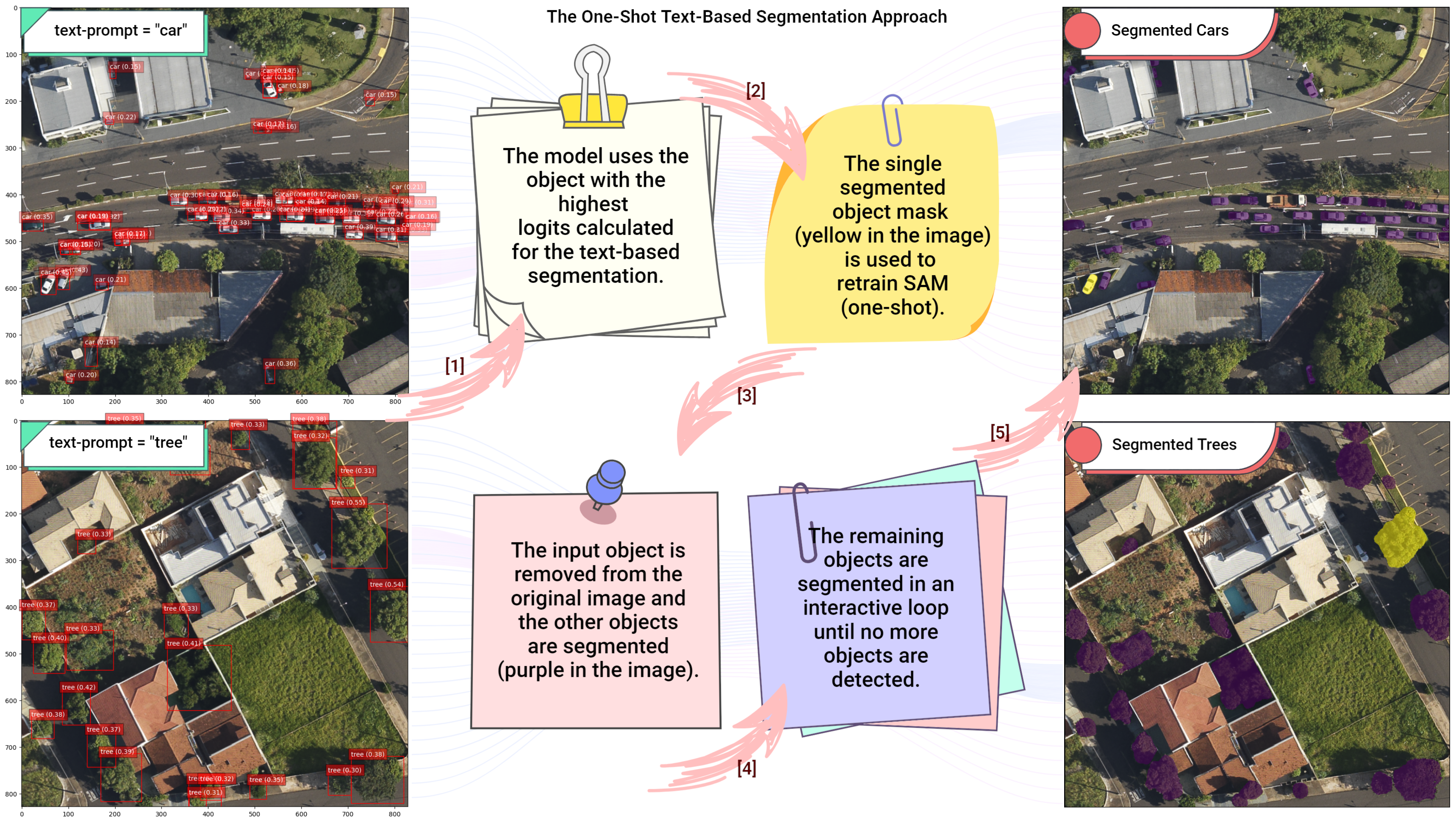}
\caption{\small \centering Visual representation of the one-shot-based text segmentation process in action. The figure provides a step-by-step illustration of how the model identifies and segments the most probable object based on a text prompt with "car" and "tree" as examples. \label{fig_oneshot}}
\end{figure*}

The one-shot training was conducted following the recommendation in \citep{zhang2023personalize} by using its PerSAM and PerSAM-F approaches. We begin by adapting the text-based approach of the combination of the GroundDINO \citep{liu2023grounding} and SAM \citep{kirillov2023segment} methods to return the overall most probable object belonging to the specified class in its description. By doing so, we enable an automated process of identifying a single object and including it on a personalized pipeline for training SAM with this novel knowledge. In this section, we describe the procedures involved in the one-shot training mechanism as well as the methods used for object identification and personalization. To summarize the whole process, we illustrate the main phases in Figure~\ref{fig_oneshot}.

Following Figure~\ref{fig_oneshot}, the initial phase of the one-shot training mechanism involves the model derived from the object with the highest logits calculated from the text-based segmentation. This ensures the object is accurately recognized and selected for further steps. It's this aspect of the process that the text-based approach starts, capitalizing on GroundDINO's capabilities for zero-shot visual grounding combined with SAM's object segmentation for pre-trained model retrieval. As such, the selected object becomes the "sample" of the one-shot training process due to its high probability of belonging to the specified class by the text.

Once the object has been identified through this method, the next phase involves creating a single-segmented object mask. This mask is used for the retraining of SAM in a one-shot manner. The text-based approach adds value by helping SAM distinguish between the different object instances present in the remote sensing imagery, such as multiple "houses", "cars", or "trees", for example. Each object is identified based on its individual likelihood, leading to the creation of a unique mask for retraining SAM. The third phase starts once the object with the highest probability has been identified and its mask has been used for SAM's one-shot training. The selected input object is removed from the original image, making the remaining objects ready for further segmentation.

The final phase involves a dynamic, interactive loop, where the remaining objects are continuously segmented until no more objects are detectable by the PerSAM approach \citep{zhang2023personalize}. This phase is critical as it ensures that every potential object within the image is identified and segmented. Here again, the loop approach aids the process, using a procedure that identifies the next highest probable object, as it creates a mask, removes it from the image, and repeats. This cycle continues until a breakpoint is reached, where it detects the previous position again.

Another important aspect of the one-shot approach regards the choice of the method for its training. An early exploration of both PerSAM and PerSAM-F methods \citep{zhang2023personalize} was conducted to assess their utility in the context of remote sensing imagery. Our investigations have shown that PerSAM-F emerges as a more suitable choice for this specific domain. PerSAM, in its original formulation, leverages one-shot data through a series of techniques such as target-guided attention, target-semantic prompting, and cascaded post-refinement, delivering favorable personalized segmentation performance for subjects in a variety of poses or contexts. However, there were occasional failure cases, notably where the subjects comprised hierarchical structures to be segmented. 

\begin{figure}[ht!]
\centering
\includegraphics[width=\columnwidth]{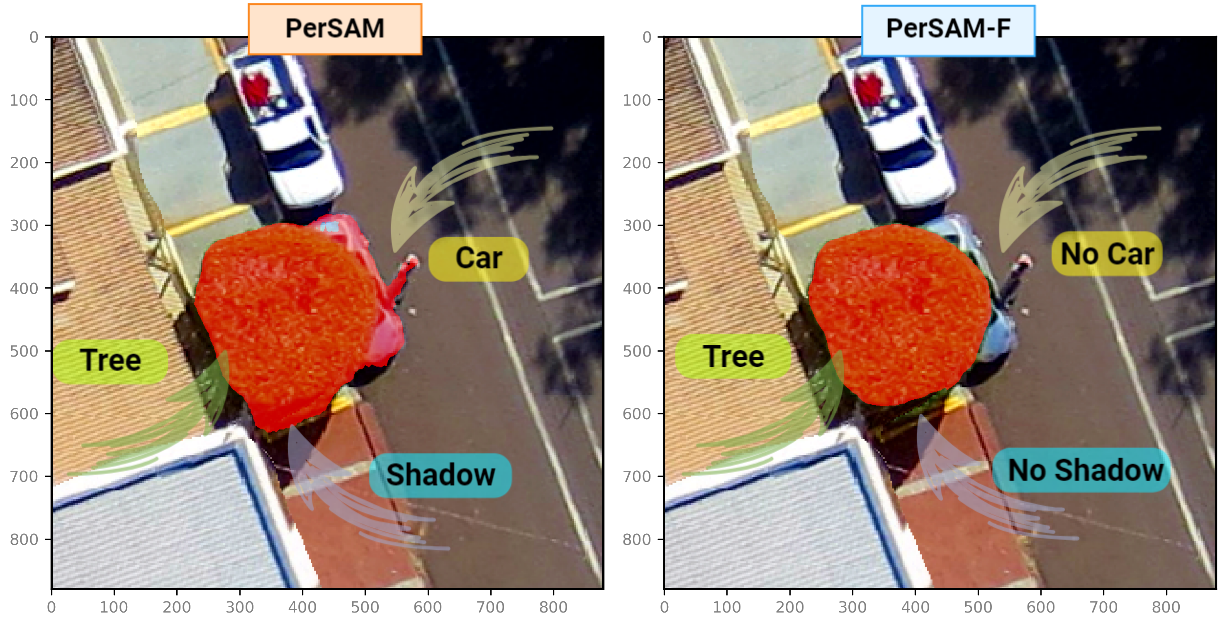}
\caption{\small \centering Comparative illustration of tree segmentation using PerSAM and PerSAM-F. On the left, the PerSAM model segments not only the tree but also its shadow and a part of the car underneath it. On the right, the PerSAM-F model, fine-tuned for hierarchical structures and varying scales, accurately segments only the tree, demonstrating its improved ability to discern and isolate the target object in remote sensing imagery.  \label{fig_freeze}}
\end{figure}

Examples of such cases in traditional images are discussed in \citep{zhang2023personalize}, where ambiguity provides a challenge for PerSAM in determining the scale of the mask as output (e.g. a "dog wearing a hat" may be segmented entirely, instead of just the "dog"). In the context of remote sensing imagery, such hierarchical structures are commonly encountered. An image may contain a tree over a house, a car near a building, a river flowing through a forest, and so forth. These hierarchical structures pose a challenge to the PerSAM method, as it struggles to determine the appropriate scale of the mask for the segmentation output. An example of such a case, where a tree covers a car, can be seen in Figure~\ref{fig_freeze}.

To address this challenge, we used PerSAM-F, the fine-tuning variant of PerSAM. As previously mentioned, PerSAM-F freezes the entire SAM to preserve its pre-trained knowledge and only fine-tunes two parameters within a ten seconds training window \citep{zhang2023personalize}. Crucially, it enables SAM to produce multiple segmentation results with different mask scales, thereby allowing for a more accurate representation of hierarchical structures commonly found in remote sensing imagery. PerSAM-F employs learnable relative weights for each scale, which adaptively select the best scale for varying objects. This strategy offers an efficient way to handle the complexity of segmentation tasks in remote sensing imagery, particularly when dealing with objects that exhibit a range of scales within a single image. This, in turn, preserves the characteristics of the segmented objects more faithfully.

As such, PerSAM-F exhibited better segmentation accuracy in our early experiments, thus being the chosen method to be incorporated with the text-based approach. In our training phase with PerSAM-F, the DICE loss and Sigmoid Focal Loss are computed, and their summation forms the final loss that is backpropagated to update the model weights. The learning rate is scheduled using the Cosine Annealing method \citep{loshchilov2017sgdr}, and the model is trained for 1000 epochs. With hardware acceleration incorporated, the model can be trained within a reasonable time frame without requiring excessive computational resources. This careful setup ensures the extraction of meaningful features from the reference image, contributing to the effectiveness of our one-shot text-based approach.

To evaluate the performance and utility of the text-based one-shot learning method, we conduct a comparative analysis against a traditional one-shot learning approach. The traditional method used for comparison follows the typical approach of one-shot learning, providing the model with a single example from the ground-truth mask, manually labeled by human experts. To ensure fairness, we provided the model with multiple random samples from each dataset, and mimic the image inputs to return a direct comparison for both approaches. We calculated the evaluation metrics from each input and returned its average value alongside with its standard deviation. Since the text approach always uses the same input (i.e. the highest logits object), we were able to return a single measurement of their accuracies.

\subsection{Model Evaluation}

The performance of both zero-shot and one-shot models was measured by evaluating their prediction accuracy on a ground-truth mask. For that, we used metrics like Intersection over Union (IoU), Pixel Accuracy, and Dice Coefficient. These metrics are commonly used in evaluating imaging segmentation, as they provide a more nuanced understanding of model performance. For that, we compared pairs of predicted and ground-truth masks.

Intersection over Union (IoU) is a common evaluation metric for object detection and segmentation problems. It measures the overlap between the predicted segmentation and the ground truth \citep{Rahman2016}. The IoU is the area of overlap divided by the area of the union of the predicted and ground truth segmentation. A higher IoU means a more accurate segmentation. The equation to achieve it is presented as:

\begin{equation}
IoU = \frac{TP}{TP + FP + FN}
\end{equation}

Here, TP represents True Positives (the correctly identified positives), FP represents False Positives (the incorrectly identified positives), and FN represents False Negatives (the positives that were missed).

Pixel Accuracy is the simplest used metric and it measures the percentage of pixels that were accurately classified \citep{Minaee2021}. It's calculated by dividing the number of correctly classified pixels by the total number of pixels. This metric can be misleading if the classes are imbalanced. The following equation returns it:

\begin{equation}
Pixel \ Accuracy = \frac{TP + TN}{TP + FP + TN + FN}
\end{equation}

Here, TN represents True Negatives (the correctly identified negatives).

Dice Coefficient (also known as the Sørensen–Dice index) is another metric used to gauge the performance of image segmentation methods. It's particularly useful for comparing the similarity of two samples. The Dice Coefficient is twice the area of overlap of the two segmentations divided by the total number of pixels in both images (the sum of the areas of both segmentations) \citep{Minaee2021}. The Dice Coefficient ranges from 0 (no overlap) to 1 (perfect overlap). The equation to perform it is given as follows:

\begin{equation}
Dice \ Coefficient = 2 * \frac{TP}{2*TP + FP + FN}
\end{equation}

We also utilized other metrics, particularly, True Positive Rate (TPR) and False Positive Rate (FPR) to measure the effectiveness of SAM, juxtaposed with the accurately labeled class from each dataset. The interpretation of these metrics as per \citep{powers2020evaluation} is: The True Positive Rate (TPR) denotes the fraction of TP cases among all actual positive instances, while the False Positive Rate (FPR) signifies the fraction of FP instances out of all negative instances. A model with a higher TPR is proficient at correctly pinpointing lines and edges and performs better at avoiding incorrect detections of lines and edges when the FPR is lower. Both metrics are calculated as:

\begin{linenomath}
\begin{equation}
\text{TPR} = \frac{TP}{(TP + FN)}
\end{equation}
\end{linenomath}

\begin{linenomath}
\begin{equation}
\text{FPR} = \frac{FP}{(FP + TN)}
\end{equation}
\end{linenomath}

In alignment with the inherent structure of SAM, a transformer network, our objective was to maintain the comprehensive context of our images to fully harness the model’s attention mechanism. This consideration led to our decision to process larger image crops or entire orthomosaics as a single unit, rather than fragmenting them into fixed-sized smaller patches. While this approach enhances the model’s contextual understanding, it understandably augments the computational time.

For most larger patches or quartered orthomosaics, the inference duration on a GPU was kept under 10 minutes, providing a balance between computational load and contextual analysis. When processing entire datasets as a whole, the time requirement extended to approximately 1 to 2 hours. Despite the augmented processing time for larger datasets, the assurance of comprehensive contextual analysis justifies this computational investment. Still, in fixed-sized patches such as the ones from the publicly available datasets, the inference time was under a second for each patch. These inferences were executed on an NVIDIA RTX 3090 equipped with 24 GB GDDR6X video memory and 10,496 CUDA cores, operating on Ubuntu 22.04.

\section{Results and Discussion}

\subsection{General Segmentation}

\begin{figure*}[ht!]
\centering
\includegraphics[width=\textwidth]{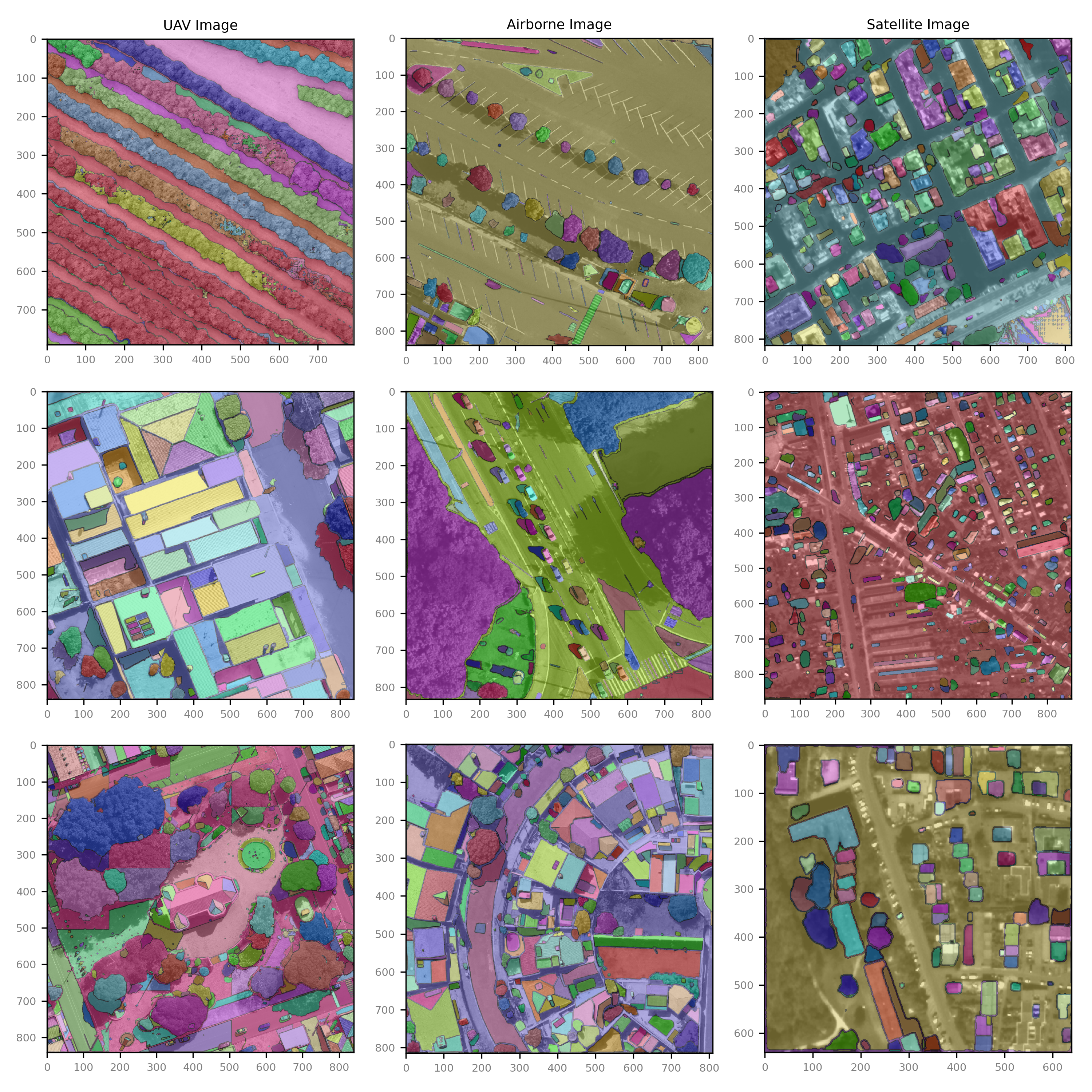}
\caption{\small \centering Examples of segmented objects using SAM's general segmentation method, drawn from diverse datasets based on their platforms. Objects are represented in random colors. As the model operates without any external inputs, it deduces object boundaries leveraging its zero-shot learning capabilities. \label{fig_zerogen}}
\end{figure*}

Our exploration of SAM for remote sensing tasks involved an evaluation of its performance across various datasets and scenarios. This section presents the results and discusses their implications for SAM's role in remote sensing image analysis. This process commenced with an investigation of SAM's general segmentation approach, which requires no prompts. By merely feeding SAM with remote sensing images, we aimed to observe its inherent ability to detect and distinguish objects on the surface. Examples of different scales are illustrated in Figure~\ref{fig_zerogen}, where we converted the individual regions to vector format. This approach demonstrates its adaptability and suitability for various applications. However, as this method is not guided by a prompt, it is not returning specific segmentation classes, making it difficult to measure its accuracy based on our available labels. 

As depicted in Figure~\ref{fig_zerogen}, the higher the spatial resolution of an image, the more accurately SAM segmented the objects. An interesting observation pertained to the processing of satellite images where SAM encountered difficulties in demarcating the boundaries between contiguous objects (like large fragments of trees or roads). Despite this limitation, SAM exhibited an ability to distinguish between different regions when considering very-high spatial resolution imagery, indicative of an effective segmentation capability that does not rely on any prompts. This approach offers value for additional applications that are based on object regions, such as classification algorithms. Moreover, SAM can expedite the process of object labeling for refining other models, thereby significantly reducing the time and manual effort required for this purpose.

\subsection{Zero-Shot Segmentation}

Following this initial evaluation, we proceeded to test SAM's promptable segmentation abilities using bounding boxes, points, and text features. The resulting metrics for each dataset are summarized in Table~\ref{tab_zero}. Having compiled a dataset across diverse platforms, including UAVs, aircraft devices, and satellites with varying pixel sizes, we noted that SAM's segmentation efficacy is also quantitatively influenced by the image's spatial resolution. These findings underscore the significant influence of spatial resolution on the effectiveness of different prompt types.

\begin{table*}[ht!]
\caption{\small \centering Summary of metrics for the image segmentation task across different platforms, targets, and resolutions, and using different prompts for SAM in zero-shot mode. The values in red indicate the best performance for a particular target under specific conditions.}
\centering
\label{tab_zero}
\resizebox{13cm}{!}{%
\begin{tabular}{cccccccccc}
\hline
\textbf{\#} & \textbf{Platform} & \textbf{Target} & \textbf{Resolution} & \textbf{Prompt} & \textbf{Dice} & \textbf{IoU} & \textbf{Pixel Acc.} & \textbf{TPR} & \textbf{FPR} \\ \hline
00 & UAV & Tree & 0.04 m & Box & 0.888 & 0.799 & 0.960 & 0.942 & 0.036 \\
 &  & & & Point & 0.918 & 0.848 & 0.976 & 0.916 & 0.014 \\
 &  & & & {\color[HTML]{C00000} Text} & {\color[HTML]{C00000} 0.922} & {\color[HTML]{C00000} 0.852} & {\color[HTML]{C00000} 0.981} & {\color[HTML]{C00000} 0.921} & {\color[HTML]{C00000} 0.012} \\
01 & UAV & House & 0.04 m & {\color[HTML]{C00000} Box} & {\color[HTML]{C00000} 0.927} & {\color[HTML]{C00000} 0.863} & {\color[HTML]{C00000} 0.984} & {\color[HTML]{C00000} 0.974} & {\color[HTML]{C00000} 0.015} \\
 &  & & & Point & 0.708 & 0.548 & 0.840 & 0.966 & 0.192 \\
 &  & & & Text & 0.892 & 0.798 & 0.956 & 0.971 & 0.101 \\
02 & UAV & Plantation & 0.01 m & Box & 0.862 & 0.828 & 0.855 & 0.882 & 0.111 \\
 &  & & & {\color[HTML]{C00000} Point} & {\color[HTML]{C00000} 0.958} & {\color[HTML]{C00000} 0.920} & {\color[HTML]{C00000} 0.950} & {\color[HTML]{C00000} 0.980} & {\color[HTML]{C00000} 0.092} \\
 &  & & & Text & 0.671 & 0.644 & 0.665 & 0.686 & 0.120 \\
03 & UAV & Plantation & 0.04 m & {\color[HTML]{C00000} Box} & {\color[HTML]{C00000} 0.801} & {\color[HTML]{C00000} 0.689} & {\color[HTML]{C00000} 0.952} & {\color[HTML]{C00000} 0.944} & {\color[HTML]{C00000} 0.104} \\
 &  & & & Point & 0.727 & 0.571 & 0.935 & 0.934 & 0.065 \\
 &  & & & Text & 0.441 & 0.328 & 0.499 & 0.450 & 0.061 \\
04 & UAV & Building & 0.09 m & {\color[HTML]{C00000} Box} & {\color[HTML]{C00000} 0.697} & {\color[HTML]{C00000} 0.535} & {\color[HTML]{C00000} 0.813} & {\color[HTML]{C00000} 0.955} & {\color[HTML]{C00000} 0.228} \\
 &  & & & Point & 0.691 & 0.528 & 0.842 & 0.911 & 0.175 \\
 &  & & & Text & 0.663 & 0.509 & 0.772 & 0.907 & 0.240 \\
05 & UAV & Car & 0.09 m & Box & 0.788 & 0.650 & 0.970 & 0.660 & 0.002 \\
 &  & & & Point & 0.900 & 0.819 & 0.991 & 0.867 & 0.003 \\
 &  & & & {\color[HTML]{C00000} Text} & {\color[HTML]{C00000} 0.927} & {\color[HTML]{C00000} 0.843} & {\color[HTML]{C00000} 0.973} & {\color[HTML]{C00000} 0.893} & {\color[HTML]{C00000} 0.001} \\
06 & Airborne & Tree & 0.20 m & Box & 0.688 & 0.524 & 0.912 & 0.844 & 0.079 \\
 &  & & & {\color[HTML]{C00000} Point} & {\color[HTML]{C00000} 0.917} & {\color[HTML]{C00000} 0.847} & {\color[HTML]{C00000} 0.935} & {\color[HTML]{C00000} 0.883} & {\color[HTML]{C00000} 0.029} \\
 &  & & & Text & 0.890 & 0.822 & 0.907 & 0.856 & 0.037 \\
07 & Airborne & Vehicle & 0.20 m & Box & 0.861 & 0.756 & 0.995 & 0.869 & 0.003 \\
 &  & & & {\color[HTML]{C00000} Point} & {\color[HTML]{C00000} 0.863} & {\color[HTML]{C00000} 0.759} & {\color[HTML]{C00000} 0.991} & {\color[HTML]{C00000} 0.785} & {\color[HTML]{C00000} 0.001} \\
 &  & & & Text & 0.846 & 0.744 & 0.971 & 0.769 & 0.002 \\
08 & Airborne & Lake & 0.45 m & Box & 0.574 & 0.403 & 0.983 & 0.988 & 0.017 \\
 &  & & & {\color[HTML]{C00000} Point} & {\color[HTML]{C00000} 0.972} & {\color[HTML]{C00000} 0.945} & {\color[HTML]{C00000} 0.999} & {\color[HTML]{C00000} 0.991} & {\color[HTML]{C00000} 0.001} \\
 &  & & & Text & 0.894 & 0.869 & 0.919 & 0.912 & 0.008 \\
09 & Satelitte & Multiclass & 0.30 m & Box & 0.391 & 0.225 & 0.945 & 0.226 & 0.004 \\
 &  & & & {\color[HTML]{C00000} Point} & {\color[HTML]{C00000} 0.823} & {\color[HTML]{C00000} 0.567} & {\color[HTML]{C00000} 0.878} & {\color[HTML]{C00000} 0.678} & {\color[HTML]{C00000} 0.037} \\
 &  & & & Text & 0.740 & 0.510 & 0.791 & 0.610 & 0.039 \\
10 & Satelitte & Multiclass & 0.50 m & Box & 0.261 & 0.150 & 0.936 & 0.151 & 0.005 \\
 &  & & & {\color[HTML]{C00000} Point} & {\color[HTML]{C00000} 0.549} & {\color[HTML]{C00000} 0.378} & {\color[HTML]{C00000} 0.870} & {\color[HTML]{C00000} 0.452} & {\color[HTML]{C00000} 0.042} \\
 &  & & & Text & 0.494 & 0.340 & 0.783 & 0.407 & 0.044 \\ \hline
\end{tabular}}
\end{table*}

For instance, on the UAV platform, text prompts showed superior performance for object segmentation tasks such as trees, with higher Dice and IoU values. However, bounding box prompts were more effective for delineating geometrically well-defined and larger objects like houses and buildings. The segmentation of plantation crops was a unique case. Point prompts performed well at a finer 0.01 m resolution for individual plants. However, as the resolution coarsened to 0.04 m and the plantation types changed, becoming denser with the plant canopy covering entire rows, bounding box prompts outperformed the others. This outcome suggests that, for certain objects, the type of input prompt can greatly influence detection and segmentation in the zero-shot approach.

With the airborne platform, point prompts were highly effective at segmenting trees and vehicles at a 0.20 m resolution. This trend continued for the segmentation of lakes at a 0.45 m resolution. It raises the question of whether the robust performance of point prompts in these scenarios is a testament to their adaptability to very high-resolution imagery or a reflection of the target object's specific characteristics. These objects primarily consist of very defined features (like cars and vehicles) or share similar characteristics (as in bodies of water).

In the context of satellite-based remote sensing imagery, point prompts proved most efficient for multi-class segmentation at the examined resolutions of 0.30 m and 0.50 m. This can be attributed to the fact that bounding box prompts tend to overshoot object boundaries, producing more false positives compared to point prompts. This finding indicates the strong ability of point prompts to manage a diverse set of objects and categories at coarser resolutions, making them a promising tool for satellite remote sensing applications. The text-based approach was found to be the least effective, primarily due to the model's difficulty in associating low-resolution objects with words. Still, it is important to notice that, from all the datasets, the satellite multiclass problem proved to be the most difficult task for the model, with generally lower metrics than the others.

\begin{figure*}[ht!]
\centering
\includegraphics[width=\textwidth]{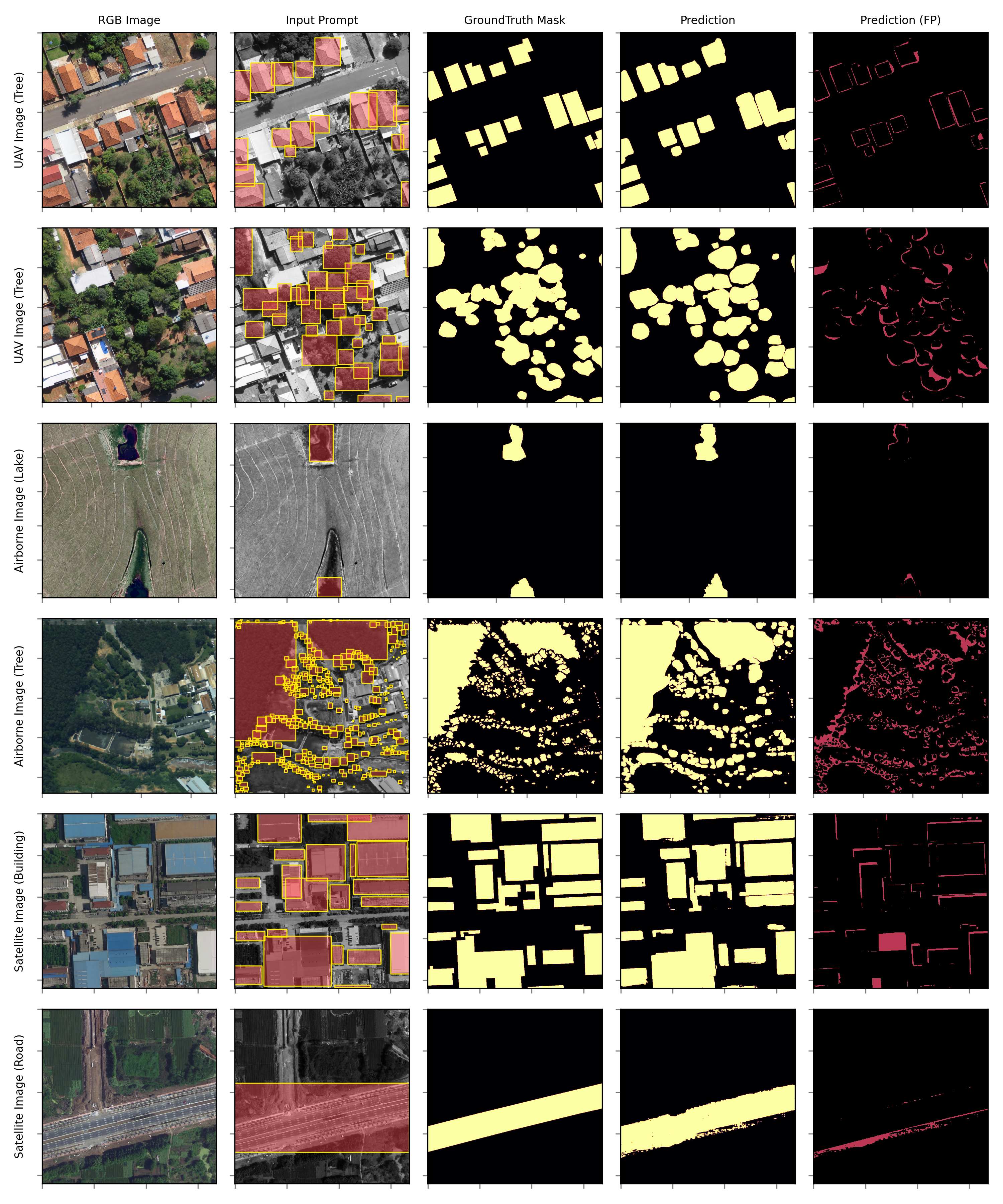}
\caption{\small \centering Illustrations of images processed using bounding-box prompts. The first column consists of the RGB image, while the second column demonstrates how the prompt was handled. The ground-truth mask is presented in the third column and the prediction result from SAM in the fourth. The last column indicates the false positive (FP) pixels from the prediction. \label{fig_zerobox}}
\end{figure*}

Qualitatively, our observations also revealed that bounding boxes were particularly effective for larger objects (Figure~\ref{fig_zerobox}). However, for smaller objects, SAM tended to overestimate the object size by including shadows in the segmented regions. Despite this overestimation, the bounding box approach still offers a useful solution for applications where an approximate estimate of such larger objects suffices. For these types of objects, a single point or central location does not suffice, they are defined by a combination of features within a particular area. Bounding boxes provide a more spatially comprehensive prompt, encapsulating the entire object, which makes them more efficient in these instances.

\begin{figure*}[ht!]
\centering
\includegraphics[width=\textwidth]{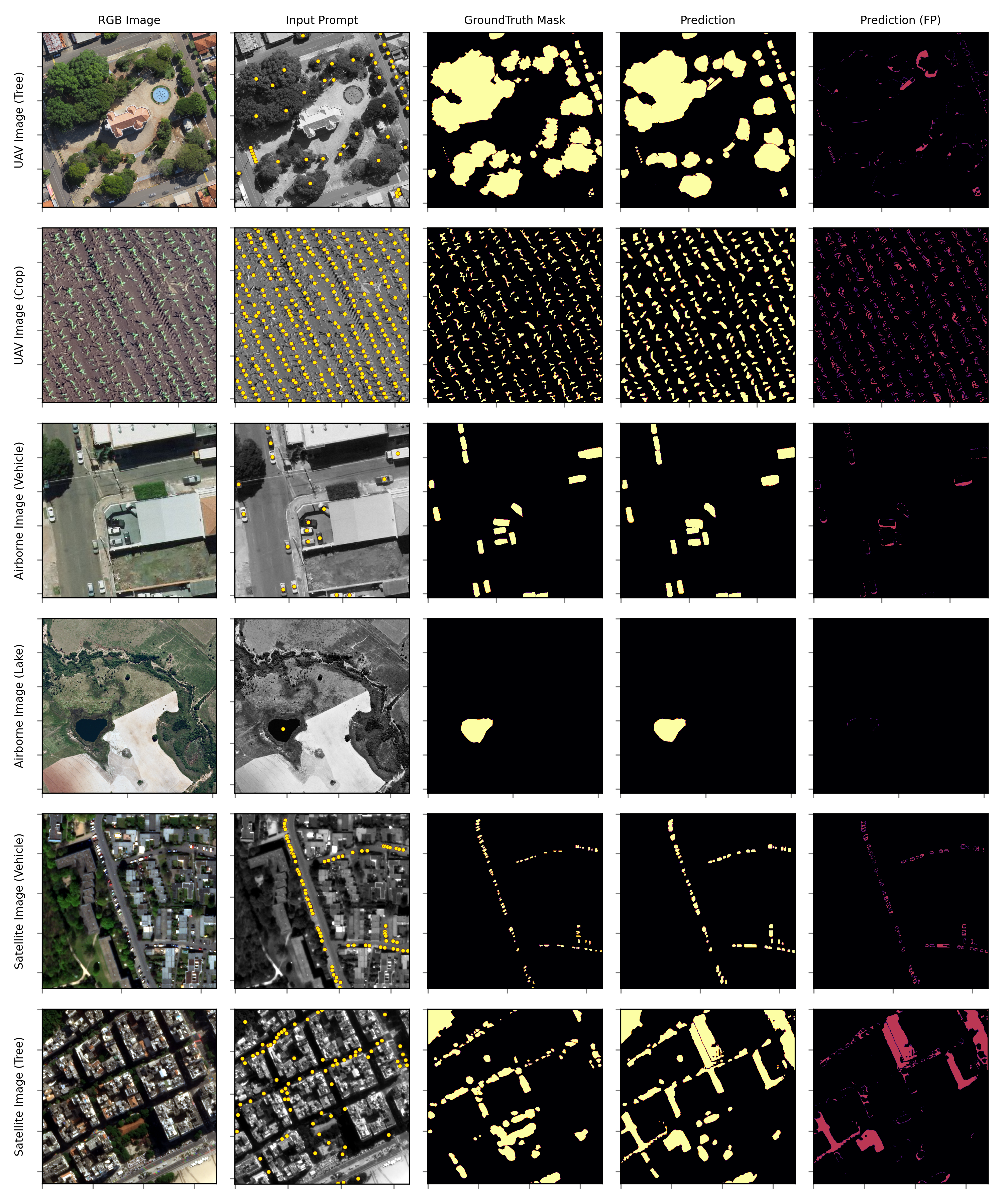}
\caption{\small \centering Illustrations of images processed using point prompts. The first column presents the RGB image, while the second column demonstrates the handling of the point prompt. The third column showcases the ground-truth mask, and the fourth column shows the prediction result from SAM. The final column highlights the false positive (FP) pixels from the prediction. \label{fig_zeropoint}}
\end{figure*}

The point-based approach outperformed the others across our dataset, specifically for distinct objects. By focusing on a singular point, SAM was able to provide precise segmentation results, thus proving its capability to work in detail (Figure~\ref{fig_zeropoint}). In the plantation dataset with 0.01 m resolution, for instance, when considering individual small plants, the point approach returned better results than bounding boxes. This approach may hold particular relevance for applications requiring precise identification and segmentation of individual objects in an image. Also, when isolating entities like single trees and vehicles, these precise spatial hints might suffice for the model to accurately identify and segment the object. 

\begin{figure*}[ht!]
\centering
\includegraphics[width=\textwidth]{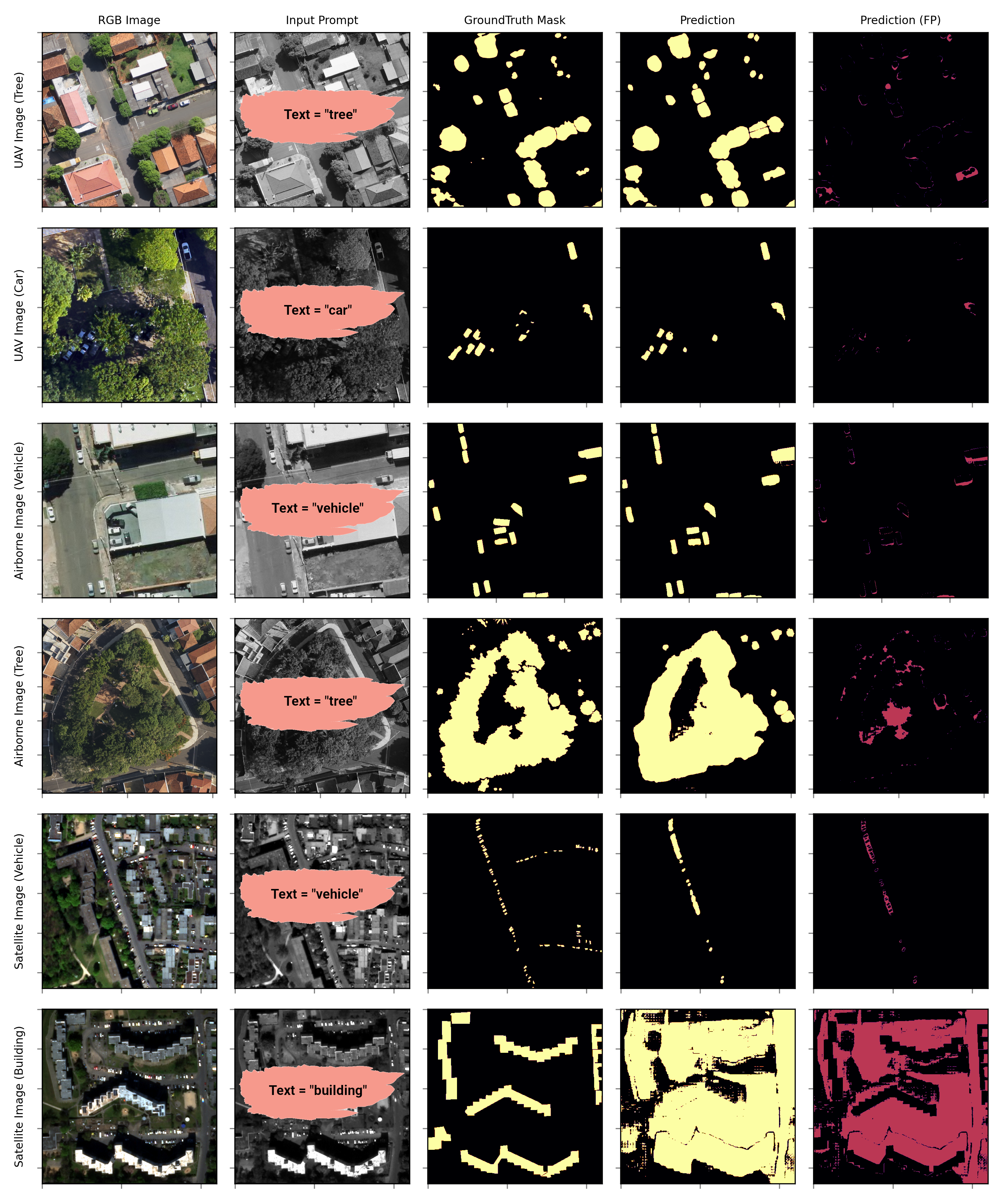}
\caption{\small \centering Examples of images processed through text-based prompts. The first column contains the RGB image, while the second column indicates the text prompt used for the model. The ground-truth mask is shown in the third column, with the prediction result from SAM in the fourth. The last column indicates the false positive (FP) pixels from the prediction. \label{fig_zerotext}}
\end{figure*}

The textual prompt approach also yielded promising results, particularly with very high-resolution images (Figure~\ref{fig_zerotext}). While it was found to be relatively comparable in performance with the point and bounding box prompts for the aerial datasets, the text prompt approach had notable limitations when used with lower spatial resolution images. The text-based approach also returned worse predictions on the plantation with 0.04 m. This may be associated with the models' limitation on understanding the characteristics of specific targets, especially when considering the bird’s eye view of remote sensing images. Since it relies on GroundDINO to interpret the text, it may be more of a limitation on it than on SAM, mostly because, when applying the general segmentation, the results visually returned overall better segmentation on these datasets (Figure~\ref{fig_zerogen}).

Text prompts, though generally trailing behind in performance, still demonstrated commendable results, often closely following the top-performing prompt type. Text prompts offer ease of implementation as their primary advantage. They don't necessitate specific spatial annotations, which are often time-consuming and resource-intensive to produce, especially for extensive remote sensing datasets. However, their effectiveness hinges on the model's ability to translate text to image information. Currently, their key limitation is that they are typically not trained specifically on remote sensing images, leading to potential inaccuracies when encountering remote sensing-specific terms or concepts. Improving the effectiveness of text prompts can be achieved through fine tuning models on remote sensing-specific datasets and terminologies. This could enable them to better interpret the nuances of remote sensing imagery, potentially enhancing their performance to match or even surpass spatial prompts like boxes and points.

\subsection{One-Shot Segmentation}

\begin{figure*}[ht!]
\centering
\includegraphics[width=\textwidth]{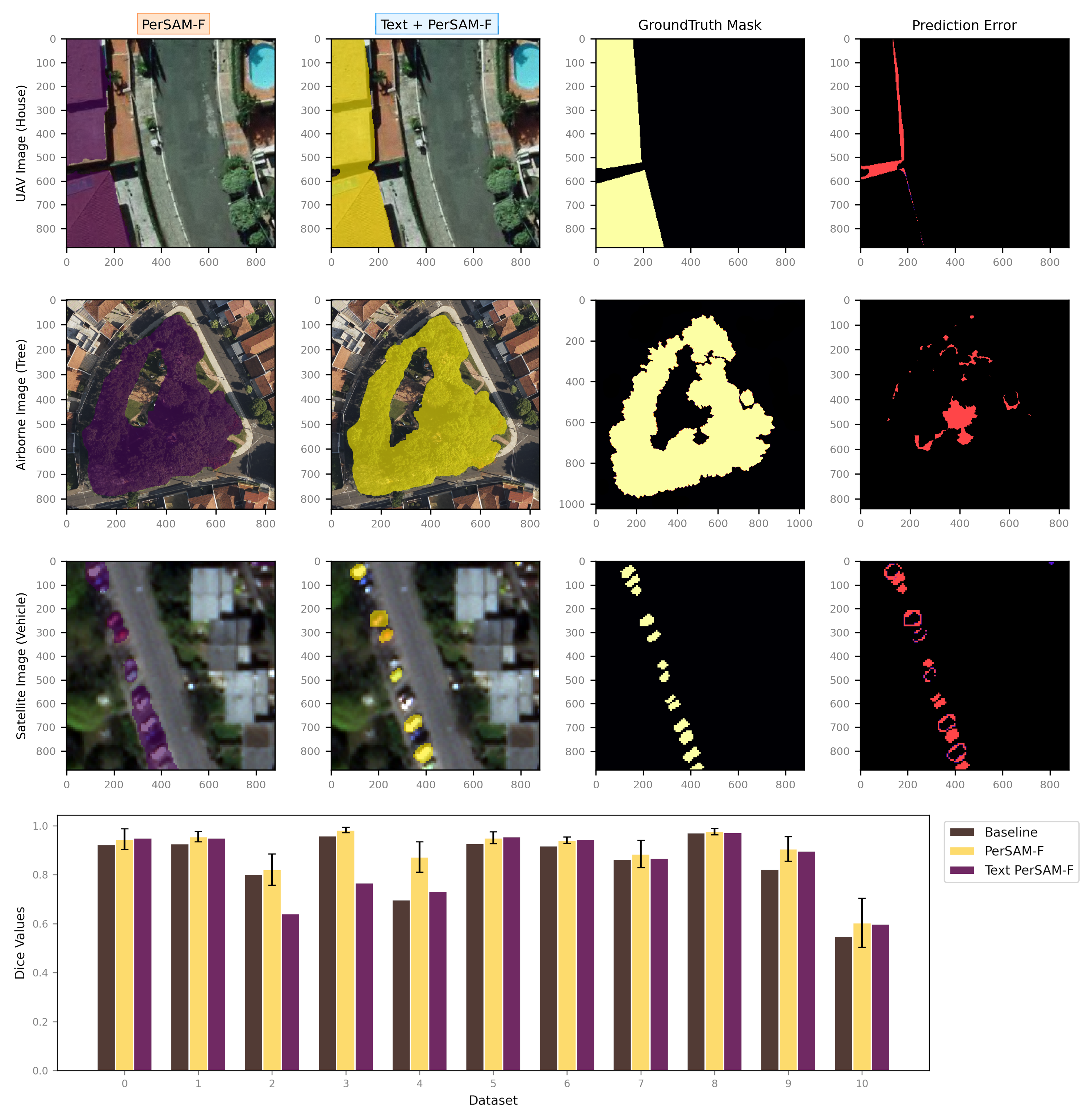}
\caption{\small \centering Visual illustration of the segmentation results using PerSAM and text-based PerSAM. from The last two columns highlights the difference in pixels the PerSAM prediction and the text-based PerSAM prediction  to its ground truth. The graphic compares the range from the Dice values of both PerSAM and text-based PerSAM, illustrating how the proposed approach remains similar to the traditional PerSAM approach, underscoring the potential of most practices to adopt the automated process in such cases. \label{fig_persam}}
\end{figure*}

Regarding our one-shot approach, we noticed that the models' performance is improved in most cases, as evidenced by the segmentation metrics calculated on each dataset. Table~\ref{tab_one} presents a detailed comparison of the different models' performance providing a summary of the segmentation results. Figure~\ref{fig_persam} offers a visual illustration of example results obtained from both approaches, particularly highlighting the performance of the model. The metrics indicate that, while the PerSAM approach with a human-sampled example may be more appropriate than the proposed text-based approach, this may not always be the case when considering the metric's standard deviation. This opens up the potential for adopting the automated process instead. However, in some instances, specifically where GroundDINO's not capable of identifying the object, to begin with, the human-labeling provides a more appropriate result.

\begin{table*}[ht!]
\caption{\small \centering Comparison of segmentation results on different platforms and targets when considering both the one-shot and the text-based one-shot approaches. The baseline values are referent to the best metric obtained by the previous zero-shot investigation, be it from a bounding box, a point, or a text prompt. The red colors indicate the best result for each scenario.}
\centering
\label{tab_one}
\resizebox{15cm}{!}{%
\begin{tabular}{cccccccccc}
\hline
\textbf{\#} & \textbf{Platform} & \textbf{Target} & \textbf{Resolution} & \textbf{Sample} & \textbf{Dice} & \textbf{IoU} & \textbf{Pixel Acc.} & \textbf{TPR} & \textbf{FPR} \\ \hline
00 & UAV & Tree & 0.04 m & Baseline & 0.922 & 0.852 & 0.981 & 0.921 & 0.012 \\
 &  & & & PerSAM-F & 0.945 ± 0.042 & 0.874 & 0.988 & 0.944 & 0.011 \\
 &  & & & {\color[HTML]{C00000} Text PerSAM-F} & {\color[HTML]{C00000} 0.950} & {\color[HTML]{C00000} 0.878} & {\color[HTML]{C00000} 0.993} & {\color[HTML]{C00000} 0.963} & {\color[HTML]{C00000} 0.009} \\
01 & UAV & House & 0.04 m & Baseline & 0.927 & 0.863 & 0.984 & 0.974 & 0.015 \\
 &  & & & {\color[HTML]{C00000} PerSAM-F} & {\color[HTML]{C00000} 0.954 ± 0.021} & {\color[HTML]{C00000} 0.889} & {\color[HTML]{C00000} 0.993} & {\color[HTML]{C00000} 0.981} & {\color[HTML]{C00000} 0.011} \\
 &  & & & Text PerSAM-F & 0.950 & 0.885 & 0.988 & 0.998 & 0.014 \\
02 & UAV & Plantation Crop & 0.01 m & Baseline & 0.801 & 0.689 & 0.952 & 0.944 & 0.104 \\
 &  & & & {\color[HTML]{C00000} PerSAM-F} & {\color[HTML]{C00000} 0.821 ± 0.064} & {\color[HTML]{C00000} 0.706} & {\color[HTML]{C00000} 0.988} & {\color[HTML]{C00000} 0.968} & {\color[HTML]{C00000} 0.096} \\
 &  & & & Text PerSAM-F & 0.641 & 0.551 & 0.762 & 0.755 & 0.156 \\
03 & UAV & Plantation Crop & 0.04 m & Baseline & 0.958 & 0.920 & 0.950 & 0.980 & 0.092 \\
 &  & & & {\color[HTML]{C00000} PerSAM-F} & {\color[HTML]{C00000} 0.982 ± 0.011} & {\color[HTML]{C00000} 0.943} & {\color[HTML]{C00000} 0.988} & {\color[HTML]{C00000} 1.004} & {\color[HTML]{C00000} 0.085} \\
 &  & & & Text PerSAM-F & 0.767 & 0.736 & 0.760 & 0.784 & 0.138 \\
04 & UAV & Building & 0.09 m & Baseline & 0.697 & 0.535 & 0.813 & 0.955 & 0.228 \\
 &  & & & {\color[HTML]{C00000} PerSAM-F} & {\color[HTML]{C00000} 0.872 ± 0.062} & {\color[HTML]{C00000} 0.669} & {\color[HTML]{C00000} 0.980} & {\color[HTML]{C00000} 0.966} & {\color[HTML]{C00000} 0.210} \\
 &  & & & Text PerSAM-F & 0.732 & 0.549 & 0.943 & 0.979 & 0.211 \\
05 & UAV & Car & 0.09 m & Baseline & 0.927 & 0.843 & 0.973 & 0.893 & 0.001 \\
 &  & & & PerSAM-F & 0.950 ± 0.024 & 0.864 & 0.988 & 0.915 & 0.001 \\
 &  & & & {\color[HTML]{C00000} Text PerSAM-F} & {\color[HTML]{C00000} 0.955} & {\color[HTML]{C00000} 0.869} & {\color[HTML]{C00000} 0.993} & {\color[HTML]{C00000} 0.933} & {\color[HTML]{C00000} 0.001} \\
06 & Airborne & Tree & 0.20 m & Baseline & 0.917 & 0.847 & 0.935 & 0.883 & 0.029 \\
 &  & & & PerSAM-F & 0.940 ± 0.013 & 0.868 & 0.988 & 0.905 & 0.027 \\
 &  & & & {\color[HTML]{C00000} Text PerSAM-F} & {\color[HTML]{C00000} 0.945} & {\color[HTML]{C00000} 0.873} & {\color[HTML]{C00000} 0.993} & {\color[HTML]{C00000} 0.923} & {\color[HTML]{C00000} 0.021} \\
07 & Airborne & Vehicle & 0.20 m & Baseline & 0.863 & 0.759 & 0.991 & 0.785 & 0.001 \\
 &  & & & {\color[HTML]{C00000} PerSAM-F} & {\color[HTML]{C00000} 0.884 ± 0.056} & {\color[HTML]{C00000} 0.778} & {\color[HTML]{C00000} 0.998} & {\color[HTML]{C00000} 0.804} & {\color[HTML]{C00000} 0.002} \\
 &  & & & Text PerSAM-F & 0.867 & 0.763 & 0.996 & 0.789 & 0.001 \\
08 & Airborne & Lake & 0.45 m & Baseline & 0.972 & 0.945 & 0.999 & 0.991 & 0.001 \\
 &  & & & {\color[HTML]{C00000} PerSAM-F} & {\color[HTML]{C00000} 0.976 ± 0.015} & {\color[HTML]{C00000} 0.949} & {\color[HTML]{C00000} 0.999} & {\color[HTML]{C00000} 0.995} & {\color[HTML]{C00000} 0.001} \\
 &  & & & Text PerSAM-F & 0.973 & 0.946 & 0.998 & 0.992 & 0.001 \\
09 & Satelitte & Multiclass & 0.30 m & Baseline & 0.823 & 0.567 & 0.878 & 0.678 & 0.037 \\
 &  & & & {\color[HTML]{C00000} PerSAM-F} & {\color[HTML]{C00000} 0.905 ± 0.052} & {\color[HTML]{C00000} 0.680} & {\color[HTML]{C00000} 0.966} & {\color[HTML]{C00000} 0.745} & {\color[HTML]{C00000} 0.035} \\
 &  & & & Text PerSAM-F & 0.897 & 0.618 & 0.958 & 0.739 & {\color[HTML]{C00000} 0.035} \\
10 & Satelitte & Multiclass & 0.50 m & Baseline & 0.549 & 0.378 & 0.870 & 0.452 & 0.042 \\
 &  & & & {\color[HTML]{C00000} PerSAM-F} & {\color[HTML]{C00000} 0.603 ± 0.104} & {\color[HTML]{C00000} 0.453} & {\color[HTML]{C00000} 0.957} & {\color[HTML]{C00000} 0.497} & {\color[HTML]{C00000} 0.039} \\
 &  & & & Text PerSAM-F & 0.598 & 0.412 & 0.948 & 0.492 & {\color[HTML]{333333} 0.040} \\ \hline
\end{tabular}}
\end{table*}

In its zero-shot form, SAM tends to favor selecting shadows in some instances alongside its target, which can lower its performance in tasks like tree detection. Segmenting objects with similar surrounding elements, especially when dealing with construction materials like streets and sidewalks, can be challenging for SAM, as noticed in our multi-class problem. Moreover, its performance with larger grouped instances, particularly when using the single-point mode, can be unsatisfactory. Also, the segmentation of smaller and irregular objects poses difficulties for SAM independently from the given prompt. SAM may generate disconnected components that do not correspond to actual features, specifically in satellite imagery where the spatial resolution is lower.

The text-based one-shot learning approach, on the other hand, automates the process of selecting the example. It uses the text-based prompt to choose the object with the highest probability (highest logits) from the image as the training example. This not only reduces the need for manual input but also ensures that the selected object is highly representative of the specified class due to its high probability. Additionally, while the text-based approach is capable of handling multiple instances of the same object class in a more streamlined manner, thanks to the looping mechanism that iteratively identifies and segments objects based on their probabilities. The one-example policy, however, excluded some of the objects in the image to favoring only the objects similar to the given sample.

In summary, upon comparing these two methods, we found that the traditional one-shot learning approach outperforms the zero-shot learning approach in all datasets. Additionally, the combination of text-based with one-shot learning also, even when not improving on it, gets close enough in most cases. This comparison underscores the benefits and potential of integrating state-of-the-art models with natural language processing capabilities for efficient and accurate geospatial analysis. Nevertheless, it is important to remember that the optimal choice between these methods may vary depending on the specific context and requirements of a given task.

\section{Future Perspectives on SAM for Remote Sensing}

SAM has several advantages that make it an attractive option for remote sensing applications. First, it offers zero-shot generalization to unfamiliar objects and images without requiring additional training \citep{kirillov2023segment}. This capability allows SAM to adapt to the diverse and dynamic nature of remote sensing data, which often consists of varying land cover types, resolutions, and imaging conditions. Second, SAM's interactive input process can significantly reduce the time and labor required for manual image segmentation. The model's ability to generate segmentation masks with minimal input, such as a text prompt, a single point, or a bounding box, accelerates the annotation process and improves the overall efficiency of remote sensing data analysis. Lastly, the decoupled architecture of SAM, comprising a one-time image encoder and a lightweight mask decoder, makes it computationally efficient. This efficiency is crucial for large-scale remote sensing applications, where processing vast amounts of data on time is of utmost importance. 

However, our study consists of an initial exploration of this model, where there's still much to be investigated. In this section, we discuss future perspectives on SAM and how it can be improved upon. Despite its potential, SAM has some limitations when applied to remote sensing imagery. One challenge is that remote sensing data often come in different formats, resolutions, and spectral bands. SAM, which has been trained primarily on RGB images, may not perform optimally with multispectral or hyperspectral data, which are common in remote sensing applications. A possible approach to this issue consists of either adapting SAM to read in multiple bands by performing rotated 3-band combinations or performing a fine-tuning to domain adaption. In our early experiments, a simple example run on different multispectral datasets demonstrated that, although the model has the potential to segment different regions or features, it still needs further exploration. This is something that we intend to explore in future research, but expect that others may look into it as well.

Regardless, the current model can be effectively used in various remote sensing applications. For instance, we verified that SAM can be easily employed for land cover mapping, where it can segment forests, urban areas, and agricultural fields. It can also be used for monitoring urban growth and land use changes, enabling policymakers and urban planners to make informed decisions based on accurate and up-to-date information. Furthermore, SAM can be applied in a pipeline process to monitor and manage natural resources. Its efficiency and speed make it suitable for real-time monitoring, providing valuable information to decision-makers. This is also a feature that could be potentially explored by research going forward with its implementation.

Nevertheless, it is crucial to underscore a significant limitation concerning the complexity of our data. While our primary objective was to analyze results across varying spatial resolutions and broad remote sensing segmentation tasks, the limited regional diversity of our data may not fully capture the range of object characteristics encountered worldwide. Future research, therefore, could emphasize utilizing and adapting to a more diverse array of the same object, thereby bolstering the robustness and applicability of the model or its adaptations. For instance, in the detection of buildings and water bodies, exploration of publicly available datasets from diverse regions \citep{boguszewski2022landcoverai, Zhang2023} could provide a more comprehensive understanding of these objects' varied characteristics, and contribute to the enhancement of algorithmic performance across varied geographical contexts.

For the one-shot technique based on SAM, which is the capacity to generate accurate segmentation from a single example \citep{zhang2023personalize}. Our experimental results indicate an improvement in performance across most investigated datasets, especially considering the border of the objects. However, it is essential to note that one-shot learning may pose challenges to the generalization capability of the model. This may be an issue of remote sensing data that often exhibit a high degree of heterogeneity and diversity \citep{Zia2022}. For instance, a "healthy" tree can be a good sample for the model, but it can bias it to ignore "unhealthy" trees or canopies with different structures.

Expanding the one-shot learning to a few-shot scenario could potentially improve the model's adaptability to different environments or tasks by enabling it to learn from more than one example (2 to 10) instead of a single one. This would involve using a small set of labeled objects for each land cover type during the training process \citep{Sun2021, Li2022}. A more robust learning approach, which uses a larger number of examples for each class, could further enhance the model's ability to capture the nuances and variations within each class. This approach, however, may require more computational resources and training data, and thus may not be suitable for all applications.

Additionally, While SAM is a powerful tool for image segmentation, its effectiveness can be boosted when combined with other techniques. For example, integrating SAM into another ViT framework in a weakly-supervised manner could potentially improve the segmentation result, better handling the spatial-contextual information. However, it's worth noting that integrating it might also bring new challenges \citep{Wang2020b}. One potential issue could be the increased model complexity and computational requirements, which might limit its feasibility. But, as the training of transformers typically requires large amounts of data, SAM can provide fast and relatively accurate labeled regions for it.

Furthermore, one of the key challenges to tackle would be improving SAM's performance when applied to low spatial resolution imagery. Thus, as the original training data of SAM primarily consisted of high-resolution images, it is inherently more suitable for similar high-resolution conditions, even in the remote sensing domain. The noticeable decrease in accuracy at resolutions above 30 cm, noted in our tests, further substantiates this observation. This shortcoming can be further explored by coupling SAM with a Super-Resolution (SR) technique \citep{Yang2015}, for instance, creating a two-step process, where the first step involves using an SR model to increase the spatial resolution of the imagery, and the second step involves using the enhanced resolution image as an input to SAM. It is acknowledged that while this method can theoretically enhance the performance of SAM with low-resolution images, the Super-Resolution techniques themselves can introduce errors, potentially offsetting the benefits \citep{Yang2015}. Therefore, the proposed two-step process should be approached with caution, ensuring meticulous testing and validation. A dedicated exploration into refining and optimizing SAM for lower-resolution images, possibly involving adaptation and training of the model on lower-resolution data, will be integral to ensuring its effective and reliable application in diverse remote sensing scenarios.

As we explored the integration of SAM with other types of methods, such as GroundDINO \citep{liu2023grounding}, we noticed both strengths and limitations that were already discussed in the previous section. This combination demonstrates a high degree of versatility and accuracy in tasks such as instance segmentation, where GroundDINO's object detection and classification guided SAM's segmentation process. However, the flexibility of this approach extends beyond these specific models. Any similar models could be swapped in as required, expanding the applications and robustness of the system. Alternatives such as GLIP \citep{li2022grounded} or CLIP \citep{liu2023remoteclip} may replace GroundDINO, allowing for further experimentation and optimization \citep{zhang2022glipv2}. Furthermore, integrating language models like ChatGPT \citep{openai2023gpt4} could offer additional layers of interaction and nuances of understanding, demonstrating the far-reaching potential of combining these expert models. This modular approach underpins a potent and adaptable workflow that could reshape our capabilities in handling remote sensing tasks.

The integration of Geographical Information Systems (GIS) with models like SAM holds significant promise for enhancing the annotation process for training specific segmentation and change detection models. A fundamental challenge often lies in the discrepancy between training data and the image data employed due to different acquisition times and since the data used could be marred with annotator errors, leading to a compatibility issue with the used image. The integration with SAM could help users optimize the creation of annotations and, when suitable, improve its results with editing, thus creating a quicker and more robust dataset. Lastly, a topic which is not discussed in this paper, but which is an important issue for applications particularly in the area of geospatial intelligence is AI security. A recent survey paper on this topic is \citep{Xu2023}. It discusses issues such as that it can be unclear based on which data a (foundation) model has been trained and what deficits may arise from this. Particularly, an adversary might have contaminated the training data.

In short, our study focused on demonstrating the potential of SAM adaptability for the remote sensing domain, as well as presenting a novel, automated approach, to retrain the model with one example from the text-based approach. While there is much to be explored, it is important to understand how the model works and how it could be improved upon. To summarize this discussion, there are many potential research directions and applications for SAM in remote sensing applications, which can be condensed as follows:
\begin{itemize}
\item Examining the most effective approaches and techniques for adapting SAM to cater to a variety of remote sensing data, including multispectral and hyperspectral data.
\item Analysing the potential of coupling SAM with few-shot or multi-shot learning, to enhance its adaptability and generalization capability across diverse remote sensing scenarios.
\item Investigating potential ways to integrate SAM with prevalent remote sensing tools and platforms, such as Geographic Information Systems (GIS), to augment the versatility and utility of these systems.
\item An issue particularly important for applications in the area of geospatial intelligence is AI security, where an adversary might, e.g., contaminate the training data for a (foundation) model.
\item Assessing the performance and efficiency of SAM in real-time or near-real-time remote sensing applications to understand its capabilities for timely data processing and analysis.
\item Exploring how domain-specific knowledge and expertise can be integrated into SAM to enhance its ability to understand and interpret remote sensing data.
\item Evaluating the potential use of SAM as an alternative to traditional labeling processes and its integration with other image classification and segmentation techniques in a weakly-supervised manner to boost its accuracy and reliability.
\item Integrating SAM with super resolution approach to enhance its capability to handle low-resolution imagery, thereby expanding the range of remote sensing imagery it can effectively analyze.
\end{itemize}

\section{Conclusions}

In this study, we conducted a comprehensive analysis of both the zero and one-shot capabilities of the Segment Anything Model (SAM) in the domain of remote sensing imagery processing, benchmarking it against aerial and satellite datasets. Our analysis provided insights into the operational performance and efficacy of SAM in the sphere of remote sensing segmentation tasks. We concluded that, while SAM exhibits notable promise, there is a tangible scope for improvement, specifically in managing its limitations and refining its performance for task-specific implementations.

In summary, our data indicated that SAM delivers notable performance when contrasted with the ground-truth masks, thereby underscoring its potential efficacy as a significant resource for remote sensing applications. Our evaluation reveals that the prompt capabilities of SAM (text, point, box, and general), combined with its ability to perform object segmentation with minimal human supervision, can also contribute to a significant reduction in annotation workload. This decrease in human input during the labeling phase may lead to expedited training schedules for other methods, thus promoting more streamlined and cost-effective workflows.

The chosen datasets were also selected with the express purpose of representing a broad and diverse context at varying scales, rather than exemplifying complex or challenging scenarios. By focusing on more straightforward datasets, the study went in on the fundamental aspects of segmentation tasks, without the additional noise of overly complicated or intricate scenarios. In this sense, future research should be oriented towards improving SAM's capabilities and exploring its potential integration with other methods to address more complex and challenging remote sensing scenarios.

Nevertheless, despite the demonstrated generalization, there are certain limitations to be addressed. Under complex scenarios, the model faces challenges, leading to less optimal segmentation outputs, by overestimating most of the objects' boundaries. Additionally, SAM's performance metrics display variability contingent on the spatial resolution of the input imagery (i.e., being prone to increase mistakes as the spatial resolution of the imagery is lowered). Consequently, identifying and rectifying these constraints is essential for further enhancing SAM's applicability within the remote sensing domain.

\section*{Supplementary}

Here, we provide an open-access repository designed to facilitate the application of the Segment Anything Model (SAM) within the domain of remote sensing imagery. The incorporated codes and packages provide users the means to implement point and bounding box-based shapefiles in combination with the SAM. The repositories also include notebooks that demonstrate how to apply the text-based prompt approach, alongside one-shot modifications of SAM. These resources aim to bolster the usability of the SAM approach in diverse remote sensing contexts, and can be accessed via the following online repositories: \href{https://github.com/LucasOsco/AI-RemoteSensing}{GitHub: AI-RemoteSensing} \citep{AIRemoteSensing2023} and; \href{https://github.com/opengeos/segment-geospatial}{GitHub: Segment-Geospatial} \citep{segment-geospatial2023}.


\section*{Acknowledgements}

This study was financed in part by the Coordenação de Aperfeiçoamento de Pessoal de Nível Superior (CAPES) - Finance Code 001. The authors are funded by the Support Foundation for the Development of Education, Science, Technology of the State of Mato Grosso do Sul (FUNDECT; 71/009.436/2022), the Brazilian National Council for Scientific and Technological Development (CNPq; 433783/2018-4, 310517/2020-6; 405997/2021-3; 308481/2022-4; 305296/2022-1), and CAPES Print (88881.311850/2018-01).

\section*{Conflicts of Interest}

The authors declare that they have no known competing financial interests or personal relationships that could have appeared to influence the work reported in this paper.

\section*{Abbreviations}{
The following abbreviations are used in this manuscript:\\

\noindent 
\begin{tabular}{@{}ll}
AI & Artificial Inteligence\\
CNNs & Convolutional Neural Networks\\
GANs & Generative Adversarial Networks\\
GIS & Geographic Information Systems\\
NLP & Natural Language Processing\\
SAM & Segment Anything Model\\
UAV & Unmanned Aerial Vehicle\\
ViT & Vision Transformer
\end{tabular}
}

\normalsize

@misc{alayrac2022flamingo,
      title={Flamingo: a Visual Language Model for Few-Shot Learning}, 
      author={Jean-Baptiste Alayrac and Jeff Donahue and Pauline Luc and Antoine Miech and Iain Barr and Yana Hasson and Karel Lenc and Arthur Mensch and Katie Millican and Malcolm Reynolds and Roman Ring and Eliza Rutherford and Serkan Cabi and Tengda Han and Zhitao Gong and Sina Samangooei and Marianne Monteiro and Jacob Menick and Sebastian Borgeaud and Andrew Brock and Aida Nematzadeh and Sahand Sharifzadeh and Mikolaj Binkowski and Ricardo Barreira and Oriol Vinyals and Andrew Zisserman and Karen Simonyan},
      year={2022},
      eprint={2204.14198},
      archivePrefix={arXiv},
      primaryClass={cs.CV}
}

@article{Aleissaee2023,
  doi = {10.3390/rs15071860},
  url = {https://doi.org/10.3390/rs15071860},
  year = {2023},
  month = mar,
  publisher = {{MDPI} {AG}},
  volume = {15},
  number = {7},
  pages = {1860},
  author = {Abdulaziz Amer Aleissaee and Amandeep Kumar and Rao Muhammad Anwer and Salman Khan and Hisham Cholakkal and Gui-Song Xia and Fahad Shahbaz Khan},
  title = {Transformers in Remote Sensing: A Survey},
  journal = {Remote Sensing}
}

@article{Amani2020,
  doi = {10.1109/jstars.2020.3021052},
  url = {https://doi.org/10.1109/jstars.2020.3021052},
  year = {2020},
  publisher = {Institute of Electrical and Electronics Engineers ({IEEE})},
  volume = {13},
  pages = {5326--5350},
  author = {Meisam Amani and Arsalan Ghorbanian and Seyed Ali Ahmadi and Mohammad Kakooei and Armin Moghimi and S. Mohammad Mirmazloumi and Sayyed Hamed Alizadeh Moghaddam and Sahel Mahdavi and Masoud Ghahremanloo and Saeid Parsian and Qiusheng Wu and Brian Brisco},
  title = {Google Earth Engine Cloud Computing Platform for Remote Sensing Big Data Applications: A Comprehensive Review},
  journal = {{IEEE} Journal of Selected Topics in Applied Earth Observations and Remote Sensing}
}

@article{Bai2022,
  doi = {10.1080/01431161.2022.2048319},
  url = {https://doi.org/10.1080/01431161.2022.2048319},
  year = {2022},
  month = mar,
  publisher = {Informa {UK} Limited},
  volume = {43},
  number = {5},
  pages = {1800--1847},
  author = {Yu Bai and Yu Zhao and Yajing Shao and Xinrong Zhang and Xuefeng Yuan},
  title = {Deep learning in different remote sensing image categories and applications: status and prospects},
  journal = {International Journal of Remote Sensing}
}

@article{BenjdiraRS2019,
  doi = {10.3390/rs11111369},
  url = {https://doi.org/10.3390/rs11111369},
  year = {2019},
  month = jun,
  publisher = {{MDPI} {AG}},
  volume = {11},
  number = {11},
  pages = {1369},
  author = {Bilel Benjdira and Yakoub Bazi and Anis Koubaa and Kais Ouni},
  title = {Unsupervised Domain Adaptation Using Generative Adversarial Networks for Semantic Segmentation of Aerial Images},
  journal = {Remote Sensing}
}

@article{Chi2016,
  doi = {10.1109/jproc.2016.2598228},
  url = {https://doi.org/10.1109/jproc.2016.2598228},
  year = {2016},
  month = nov,
  publisher = {Institute of Electrical and Electronics Engineers ({IEEE})},
  volume = {104},
  number = {11},
  pages = {2207--2219},
  author = {Mingmin Chi and Antonio Plaza and Jon Atli Benediktsson and Zhongyi Sun and Jinsheng Shen and Yangyong Zhu},
  title = {Big Data for Remote Sensing: Challenges and Opportunities},
  journal = {Proceedings of the {IEEE}}
}

@article{deCarvalho2022,
  doi = {10.3390/rs14040965},
  url = {https://doi.org/10.3390/rs14040965},
  year = {2022},
  month = feb,
  publisher = {{MDPI} {AG}},
  volume = {14},
  number = {4},
  pages = {965},
  author = {Osmar Luiz Ferreira de Carvalho and Osmar Ab{\'{\i}}lio de Carvalho J{\'{u}}nior and Cristiano Rosa e Silva and Anesmar Olino de Albuquerque and Nickolas Castro Santana and Dibio Leandro Borges and Roberto Arnaldo Trancoso Gomes and Renato Fontes Guimar{\~{a}}es},
  title = {Panoptic Segmentation Meets Remote Sensing},
  journal = {Remote Sensing}
}

@article{dingAIDA2021,
	doi = {10.1109/tpami.2021.3117983},
  
	url = {https://doi.org/10.1109
  
	year = 2022,
	month = {nov},
  
	publisher = {Institute of Electrical and Electronics Engineers ({IEEE})},
  
	volume = {44},
  
	number = {11},
  
	pages = {7778--7796},
  
	author = {Jian Ding and Nan Xue and Gui-Song Xia and Xiang Bai and Wen Yang and Michael Ying Yang and Serge Belongie and Jiebo Luo and Mihai Datcu and Marcello Pelillo and Liangpei Zhang},
  
	title = {Object Detection in Aerial Images: A Large-Scale Benchmark and Challenges},
  
	journal = {{IEEE} Transactions on Pattern Analysis and Machine Intelligence}
}

@online{esa2023,
    author = {{European Space Agency}},
    title = {{SkySat - EOGateway}},
    year = {2023},
    url = {https://earth.esa.int/eogateway/missions/SkySat},
    urldate = {2023-05-29}
}

@article{Gao2021,
author = {Gao, Kyle and Chen, Mengge and Narges Fatholahi, Sarah and He, Hongjie and Xu, Hongzhang and Marcato Junior, José and Nunes Gonçalves, Wesley and Chapman, Michael A. and Li, Jonathan},
title = {A region-based deep learning approach to instance segmentation of aerial orthoimagery for building rooftop extraction},
journal = {Geomatica},
volume = {75},
number = {3},
pages = {148-164},
year = {2021},
doi = {10.1139/geomat-2021-0009},

URL = { 
    
    
        https://cdnsciencepub.com/doi/abs/10.1139/geomat-2021-0009
    

},
eprint = { 
    
    
        https://cdnsciencepub.com/doi/pdf/10.1139/geomat-2021-0009
    

}
,
    abstract = { Updated building information plays an important role in many fields such as environmental monitoring, disaster assessment, and the creation of base maps for urban planning. High-resolution images captured from Earth observation satellites and airborne platforms provide valuable data that cover large areas at high temporal frequencies. In recent years, deep neural networks have shown great potential in the semantic segmentation of Earth observation images for building detection, significantly exceeding the performance of traditional machine learning methods whenever high-quality training datasets are available. Instance segmentation methods further leverage object detection to focus segmentation onto regions of interest, avoiding certain types of false positives and false negatives when compared to semantic segmentation methods. In this study, we approach building rooftop detection as an instance segmentation problem and propose a region-based deep learning approach to building rooftop extraction based on the Mask Regional Convolutional Neural Network (Mask R-CNN) framework. Our study indicates that searching for suitable hyperparameters results in considerable improvements in deep learning models. We found that hyperparameter optimization could be mandatory in some cases because in our experiments, the baseline Mask R-CNN achieved an unacceptable performance when compared to other methods. Our optimized Mask R-CNN, on the other hand, achieves a precision, recall, and F1-score of 92\%, 86.6\%, and 89.1\%, respectively. Furthermore, we show that by using a region-based instance segmentation model, we can avoid the speckle-like errors sometimes found in semantic segmentation models, resulting in clean and accurate rooftop extraction that is more suited for practical applications. }
}

@article{Gharibbafghi2018,
  doi = {10.3390/rs10111824},
  url = {https://doi.org/10.3390/rs10111824},
  year = {2018},
  month = nov,
  publisher = {{MDPI} {AG}},
  volume = {10},
  number = {11},
  pages = {1824},
  author = {Zeinab Gharibbafghi and Jiaojiao Tian and Peter Reinartz},
  title = {Modified Superpixel Segmentation for Digital Surface Model Refinement and Building Extraction from Satellite Stereo Imagery},
  journal = {Remote Sensing}
}

@article{Gmez2016,
  doi = {10.1016/j.isprsjprs.2016.03.008},
  url = {https://doi.org/10.1016/j.isprsjprs.2016.03.008},
  year = {2016},
  month = jun,
  publisher = {Elsevier {BV}},
  volume = {116},
  pages = {55--72},
  author = {Cristina G{\'{o}}mez and Joanne C. White and Michael A. Wulder},
  title = {Optical remotely sensed time series data for land cover classification: A review},
  journal = {{ISPRS} Journal of Photogrammetry and Remote Sensing}
}

@article{Gonalves2023,
  doi = {10.1016/j.jag.2022.103151},
  url = {https://doi.org/10.1016/j.jag.2022.103151},
  year = {2023},
  month = feb,
  publisher = {Elsevier {BV}},
  volume = {116},
  pages = {103151},
  author = {Diogo Nunes Gon{\c{c}}alves and Jos{\'{e}} Marcato and Andr{\'{e}} Caceres Carrilho and Plabiany Rodrigo Acosta and Ana Paula Marques Ramos and Felipe David Georges Gomes and Lucas Prado Osco and Maxwell da Rosa Oliveira and Jos{\'{e}} Augusto Correa Martins and Geraldo Alves Damasceno and M{\'{a}}rcio Santos de Ara{\'{u}}jo and Jonathan Li and F{\'{a}}bio Roque and Leonardo de Faria Peres and Wesley Nunes Gon{\c{c}}alves and Renata Libonati},
  title = {Transformers for mapping burned areas in Brazilian Pantanal and Amazon with {PlanetScope} imagery},
  journal = {International Journal of Applied Earth Observation and Geoinformation}
}

@online{groundedsam2023,
    author = {{IDEA-Research}},
    title = {{Grounded-Segment-Anything}},
    year = {2023},
    organization = {{GitHub}},
    url = {https://github.com/IDEA-Research/Grounded-Segment-Anything}
}

@article{Hossain2019,
  doi = {10.1016/j.isprsjprs.2019.02.009},
  url = {https://doi.org/10.1016/j.isprsjprs.2019.02.009},
  year = {2019},
  month = apr,
  publisher = {Elsevier {BV}},
  volume = {150},
  pages = {115--134},
  author = {Mohammad D. Hossain and Dongmei Chen},
  title = {Segmentation for Object-Based Image Analysis ({OBIA}): A review of algorithms and challenges from remote sensing perspective},
  journal = {{ISPRS} Journal of Photogrammetry and Remote Sensing}
}

@article{Hua2021,
  doi = {10.1016/j.asoc.2021.107515},
  url = {https://doi.org/10.1016/j.asoc.2021.107515},
  year = {2021},
  month = sep,
  publisher = {Elsevier {BV}},
  volume = {109},
  pages = {107515},
  author = {Xia Hua and Xinqing Wang and Ting Rui and Faming Shao and Dong Wang},
  title = {Cascaded panoptic segmentation method for high resolution remote sensing image},
  journal = {Applied Soft Computing}
}

@article{hua2021sparse,
  doi = {10.1109/lgrs.2021.3051053},
  url = {https://doi.org/10.1109/lgrs.2021.3051053},
  year = {2022},
  publisher = {Institute of Electrical and Electronics Engineers ({IEEE})},
  volume = {19},
  pages = {1--5},
  author = {Yuansheng Hua and Diego Marcos and Lichao Mou and Xiao Xiang Zhu and Devis Tuia},
  title = {Semantic Segmentation of Remote Sensing Images With Sparse Annotations},
  journal = {{IEEE} Geoscience and Remote Sensing Letters}
}

@article{Jozdani2022,
  doi = {10.1016/j.jag.2022.102734},
  url = {https://doi.org/10.1016/j.jag.2022.102734},
  year = {2022},
  month = apr,
  publisher = {Elsevier {BV}},
  volume = {108},
  pages = {102734},
  author = {Shahab Jozdani and Dongmei Chen and Darren Pouliot and Brian Alan Johnson},
  title = {A review and meta-analysis of Generative Adversarial Networks and their applications in remote sensing},
  journal = {International Journal of Applied Earth Observation and Geoinformation}
}

@misc{kirillov2023segment,
      title={Segment Anything}, 
      author={Alexander Kirillov and Eric Mintun and Nikhila Ravi and Hanzi Mao and Chloe Rolland and Laura Gustafson and Tete Xiao and Spencer Whitehead and Alexander C. Berg and Wan-Yen Lo and Piotr Dollár and Ross Girshick},
      year={2023},
      eprint={2304.02643},
      archivePrefix={arXiv},
      primaryClass={cs.CV}
}

@article{Kotaridis2021,
  doi = {10.1016/j.isprsjprs.2021.01.020},
  url = {https://doi.org/10.1016/j.isprsjprs.2021.01.020},
  year = {2021},
  month = mar,
  publisher = {Elsevier {BV}},
  volume = {173},
  pages = {309--322},
  author = {Ioannis Kotaridis and Maria Lazaridou},
  title = {Remote sensing image segmentation advances: A meta-analysis},
  journal = {{ISPRS} Journal of Photogrammetry and Remote Sensing}
}

@article{Li2020,
  doi = {10.3390/rs12050789},
  url = {https://doi.org/10.3390/rs12050789},
  year = {2020},
  month = mar,
  publisher = {{MDPI} {AG}},
  volume = {12},
  number = {5},
  pages = {789},
  author = {Kun Li and Xiangyun Hu and Huiwei Jiang and Zhen Shu and Mi Zhang},
  title = {Attention-Guided Multi-Scale Segmentation Neural Network for Interactive Extraction of Region Objects from High-Resolution Satellite Imagery},
  journal = {Remote Sensing}
}

@misc{li2022grounded,
      title={Grounded Language-Image Pre-training}, 
      author={Liunian Harold Li and Pengchuan Zhang and Haotian Zhang and Jianwei Yang and Chunyuan Li and Yiwu Zhong and Lijuan Wang and Lu Yuan and Lei Zhang and Jenq-Neng Hwang and Kai-Wei Chang and Jianfeng Gao},
      year={2022},
      eprint={2112.03857},
      archivePrefix={arXiv},
      primaryClass={cs.CV}
}

@misc{li2023uniformer,
      title={UniFormer: Unifying Convolution and Self-attention for Visual Recognition}, 
      author={Kunchang Li and Yali Wang and Junhao Zhang and Peng Gao and Guanglu Song and Yu Liu and Hongsheng Li and Yu Qiao},
      year={2023},
      eprint={2201.09450},
      archivePrefix={arXiv},
      primaryClass={cs.CV}
}

@article{Li2022,
  doi = {10.1109/tgrs.2021.3051383},
  url = {https://doi.org/10.1109/tgrs.2021.3051383},
  year = {2022},
  publisher = {Institute of Electrical and Electronics Engineers ({IEEE})},
  volume = {60},
  pages = {1--14},
  author = {Xiang Li and Jingyu Deng and Yi Fang},
  title = {Few-Shot Object Detection on Remote Sensing Images},
  journal = {{IEEE} Transactions on Geoscience and Remote Sensing}
}

@misc{li2023transformerbased,
      title={Transformer-Based Visual Segmentation: A Survey}, 
      author={Xiangtai Li and Henghui Ding and Wenwei Zhang and Haobo Yuan and Jiangmiao Pang and Guangliang Cheng and Kai Chen and Ziwei Liu and Chen Change Loy},
      year={2023},
      eprint={2304.09854},
      archivePrefix={arXiv},
      primaryClass={cs.CV}
}

@misc{liu2023grounding,
      title={Grounding DINO: Marrying DINO with Grounded Pre-Training for Open-Set Object Detection}, 
      author={Shilong Liu and Zhaoyang Zeng and Tianhe Ren and Feng Li and Hao Zhang and Jie Yang and Chunyuan Li and Jianwei Yang and Hang Su and Jun Zhu and Lei Zhang},
      year={2023},
      eprint={2303.05499},
      archivePrefix={arXiv},
      primaryClass={cs.CV}
}

@misc{liu2023remoteclip,
      title={RemoteCLIP: A Vision Language Foundation Model for Remote Sensing}, 
      author={Fan Liu and Delong Chen and Zhangqingyun Guan and Xiaocong Zhou and Jiale Zhu and Jun Zhou},
      year={2023},
      eprint={2306.11029},
      archivePrefix={arXiv},
      primaryClass={cs.CV}
}

@article{LoboTorres2020,
  doi = {10.3390/s20020563},
  url = {https://doi.org/10.3390/s20020563},
  year = {2020},
  month = jan,
  publisher = {{MDPI} {AG}},
  volume = {20},
  number = {2},
  pages = {563},
  author = {Daliana Lobo Torres and Raul Queiroz Feitosa and Patrick Nigri Happ and Laura Elena Cu{\'{e}} La Rosa and Jos{\'{e}} Marcato Junior and Jos{\'{e}} Martins and Patrik Ol{\~{a}} Bressan and Wesley Nunes Gon{\c{c}}alves and Veraldo Liesenberg},
  title = {Applying Fully Convolutional Architectures for Semantic Segmentation of a Single Tree Species in Urban Environment on High Resolution {UAV} Optical Imagery},
  journal = {Sensors}
}

@article{Lobry2020,
  doi = {10.1109/tgrs.2020.2988782},
  url = {https://doi.org/10.1109/tgrs.2020.2988782},
  year = {2020},
  month = dec,
  publisher = {Institute of Electrical and Electronics Engineers ({IEEE})},
  volume = {58},
  number = {12},
  pages = {8555--8566},
  author = {Sylvain Lobry and Diego Marcos and Jesse Murray and Devis Tuia},
  title = {{RSVQA}: Visual Question Answering for Remote Sensing Data},
  journal = {{IEEE} Transactions on Geoscience and Remote Sensing}
}

@article{Lu2021,
  doi = {10.3390/rs13091657},
  url = {https://doi.org/10.3390/rs13091657},
  year = {2021},
  month = apr,
  publisher = {{MDPI} {AG}},
  volume = {13},
  number = {9},
  pages = {1657},
  author = {Junyan Lu and Hongguang Jia and Tie Li and Zhuqiang Li and Jingyu Ma and Ruifei Zhu},
  title = {An Instance Segmentation Based Framework for Large-Sized High-Resolution Remote Sensing Images Registration},
  journal = {Remote Sensing}
}

@article{Ma2022,
  doi = {10.1109/tgrs.2021.3097148},
  url = {https://doi.org/10.1109/tgrs.2021.3097148},
  year = {2022},
  publisher = {Institute of Electrical and Electronics Engineers ({IEEE})},
  volume = {60},
  pages = {1--16},
  author = {Ailong Ma and Junjue Wang and Yanfei Zhong and Zhuo Zheng},
  title = {{FactSeg}: Foreground Activation-Driven Small Object Semantic Segmentation in Large-Scale Remote Sensing Imagery},
  journal = {{IEEE} Transactions on Geoscience and Remote Sensing}
}

@misc{mai2023opportunities,
      title={On the Opportunities and Challenges of Foundation Models for Geospatial Artificial Intelligence}, 
      author={Gengchen Mai and Weiming Huang and Jin Sun and Suhang Song and Deepak Mishra and Ninghao Liu and Song Gao and Tianming Liu and Gao Cong and Yingjie Hu and Chris Cundy and Ziyuan Li and Rui Zhu and Ni Lao},
      year={2023},
      eprint={2304.06798},
      archivePrefix={arXiv},
      primaryClass={cs.AI}
}

@article{Martins2021,
  doi = {10.3390/rs13163054},
  url = {https://doi.org/10.3390/rs13163054},
  year = {2021},
  month = aug,
  publisher = {{MDPI} {AG}},
  volume = {13},
  number = {16},
  pages = {3054},
  author = {Jos{\'{e}} Augusto Correa Martins and Keiller Nogueira and Lucas Prado Osco and Felipe David Georges Gomes and Danielle Elis Garcia Furuya and Wesley Nunes Gon{\c{c}}alves and Diego Andr{\'{e}} Sant'Ana and Ana Paula Marques Ramos and Veraldo Liesenberg and Jefersson Alex dos Santos and Paulo Tarso Sanches de Oliveira and Jos{\'{e}} Marcato Junior},
  title = {Semantic Segmentation of Tree-Canopy in Urban Environment with Pixel-Wise Deep Learning},
  journal = {Remote Sensing}
}

@misc{mialon2023augmented,
      title={Augmented Language Models: a Survey}, 
      author={Grégoire Mialon and Roberto Dessì and Maria Lomeli and Christoforos Nalmpantis and Ram Pasunuru and Roberta Raileanu and Baptiste Rozière and Timo Schick and Jane Dwivedi-Yu and Asli Celikyilmaz and Edouard Grave and Yann LeCun and Thomas Scialom},
      year={2023},
      eprint={2302.07842},
      archivePrefix={arXiv},
      primaryClass={cs.CL}
}

@article{Minaee2021,
  doi = {10.1109/tpami.2021.3059968},
  url = {https://doi.org/10.1109/tpami.2021.3059968},
  year = {2021},
  publisher = {Institute of Electrical and Electronics Engineers ({IEEE})},
  pages = {1--1},
  author = {Shervin Minaee and Yuri Y. Boykov and Fatih Porikli and Antonio J Plaza and Nasser Kehtarnavaz and Demetri Terzopoulos},
  title = {Image Segmentation Using Deep Learning: A Survey},
  journal = {{IEEE} Transactions on Pattern Analysis and Machine Intelligence}
}

@misc{openai2023gpt4,
      title={GPT-4 Technical Report}, 
      author={OpenAI},
      year={2023},
      eprint={2303.08774},
      archivePrefix={arXiv},
      primaryClass={cs.CL}
}

@article{Osco2020,
  doi = {10.1016/j.isprsjprs.2019.12.010},
  url = {https://doi.org/10.1016/j.isprsjprs.2019.12.010},
  year = {2020},
  month = feb,
  publisher = {Elsevier {BV}},
  volume = {160},
  pages = {97--106},
  author = {Lucas Prado Osco and Mauro dos Santos de Arruda and Jos{\'{e}} Marcato Junior and Neemias Buceli da Silva and Ana Paula Marques Ramos and {\'{E}}rika Akemi Saito Moryia and Nilton Nobuhiro Imai and Danillo Roberto Pereira and Jos{\'{e}} Eduardo Creste and Edson Takashi Matsubara and Jonathan Li and Wesley Nunes Gon{\c{c}}alves},
  title = {A convolutional neural network approach for counting and geolocating citrus-trees in {UAV} multispectral imagery},
  journal = {{ISPRS} Journal of Photogrammetry and Remote Sensing}
}

@article{Osco2021_cnn,
  doi = {10.1016/j.isprsjprs.2021.01.024},
  url = {https://doi.org/10.1016/j.isprsjprs.2021.01.024},
  year = {2021},
  month = apr,
  publisher = {Elsevier {BV}},
  volume = {174},
  pages = {1--17},
  author = {Lucas Prado Osco and Mauro dos Santos de Arruda and Diogo Nunes Gon{\c{c}}alves and Alexandre Dias and Juliana Batistoti and Mauricio de Souza and Felipe David Georges Gomes and Ana Paula Marques Ramos and L{\'{u}}cio Andr{\'{e}} de Castro Jorge and Veraldo Liesenberg and Jonathan Li and Lingfei Ma and Jos{\'{e}} Marcato and Wesley Nunes Gon{\c{c}}alves},
  title = {A {CNN} approach to simultaneously count plants and detect plantation-rows from {UAV} imagery},
  journal = {{ISPRS} Journal of Photogrammetry and Remote Sensing}
}

@article{Osco2021,
  doi = {10.1016/j.jag.2021.102456},
  url = {https://doi.org/10.1016/j.jag.2021.102456},
  year = {2021},
  month = oct,
  publisher = {Elsevier {BV}},
  volume = {102},
  pages = {102456},
  author = {Lucas Prado Osco and Jos{\'{e}} Marcato Junior and Ana Paula Marques Ramos and L{\'{u}}cio Andr{\'{e}} de Castro Jorge and Sarah Narges Fatholahi and Jonathan de Andrade Silva and Edson Takashi Matsubara and Hemerson Pistori and Wesley Nunes Gon{\c{c}}alves and Jonathan Li},
  title = {A review on deep learning in {UAV} remote sensing},
  journal = {International Journal of Applied Earth Observation and Geoinformation}
}

@article{Osco2021,
  doi = {10.1007/s11119-020-09777-5},
  url = {https://doi.org/10.1007/s11119-020-09777-5},
  year = {2021},
  month = jan,
  publisher = {Springer Science and Business Media {LLC}},
  volume = {22},
  number = {4},
  pages = {1171--1188},
  author = {Lucas Prado Osco and Keiller Nogueira and Ana Paula Marques Ramos and Mayara Maezano Faita Pinheiro and Danielle Elis Garcia Furuya and Wesley Nunes Gon{\c{c}}alves and Lucio Andr{\'{e}} de Castro Jorge and Jos{\'{e}} Marcato Junior and Jefersson Alex dos Santos},
  title = {Semantic segmentation of citrus-orchard using deep neural networks and multispectral {UAV}-based imagery},
  journal = {Precision Agriculture}
}

@article{Osco2023,
  doi = {10.3390/rs15133232},
  url = {https://doi.org/10.3390/rs15133232},
  year = {2023},
  month = jun,
  publisher = {{MDPI} {AG}},
  volume = {15},
  number = {13},
  pages = {3232},
  author = {Lucas Prado Osco and Eduardo Lopes de Lemos and Wesley Nunes Gon{\c{c}}alves and Ana Paula Marques Ramos and Jos{\'{e}} Marcato Junior},
  title = {The Potential of Visual {ChatGPT} for Remote Sensing},
  journal = {Remote Sensing}
}

@software{AIRemoteSensing2023,
  author       = {Lucas Osco},
  title        = {{AI-RemoteSensing: a collection of Jupyter and 
                   Google Colaboratory notebooks dedicated to
                   leveraging Artificial Intelligence (AI) in Remote
                   Sensing applications}},
  month        = jun,
  year         = 2023,
  publisher    = {Zenodo},
  version      = {educational},
  doi          = {10.5281/zenodo.8092269},
  url          = {https://doi.org/10.5281/zenodo.8092269}
}

@misc{powers2020evaluation,
      title={Evaluation: from precision, recall and F-measure to ROC, informedness, markedness and correlation}, 
      author={David M. W. Powers},
      year={2020},
      eprint={2010.16061},
      archivePrefix={arXiv},
      primaryClass={cs.LG}
}

@article{Rahnemoonfar2021,
  doi = {10.1109/access.2021.3090981},
  url = {https://doi.org/10.1109/access.2021.3090981},
  year = {2021},
  publisher = {Institute of Electrical and Electronics Engineers ({IEEE})},
  volume = {9},
  pages = {89644--89654},
  author = {Maryam Rahnemoonfar and Tashnim Chowdhury and Argho Sarkar and Debvrat Varshney and Masoud Yari and Robin Roberson Murphy},
  title = {{FloodNet}: A High Resolution Aerial Imagery Dataset for Post Flood Scene Understanding},
  journal = {{IEEE} Access}
}

@incollection{Rahman2016,
  doi = {10.1007/978-3-319-50835-1_22},
  url = {https://doi.org/10.1007/978-3-319-50835-1_22},
  year = {2016},
  publisher = {Springer International Publishing},
  pages = {234--244},
  author = {Md Atiqur Rahman and Yang Wang},
  title = {Optimizing Intersection-Over-Union in Deep Neural Networks for Image Segmentation},
  booktitle = {Advances in Visual Computing}
}

@misc{https://doi.org/10.5281/zenodo.7966658,
  doi = {10.5281/ZENODO.7966658},
  url = {https://zenodo.org/record/7966658},
  author = {Wu,  Qiusheng and Osco,  Lucas Prado},
  keywords = {Geospatial,  Segment Anything Model,  Deep learning},
  title = {samgeo: A Python package for segmenting geospatial data with the Segment Anything Model (SAM)},
  publisher = {Zenodo},
  year = {2023},
  copyright = {MIT License}
}

@inproceedings{Su2019,
  doi = {10.1109/igarss.2019.8898573},
  url = {https://doi.org/10.1109/igarss.2019.8898573},
  year = {2019},
  month = jul,
  publisher = {{IEEE}},
  author = {Hao Su and Shunjun Wei and Min Yan and Chen Wang and Jun Shi and Xiaoling Zhang},
  title = {Object Detection and Instance Segmentation in Remote Sensing Imagery Based on Precise Mask R-{CNN}},
  booktitle = {{IGARSS} 2019 - 2019 {IEEE} International Geoscience and Remote Sensing Symposium}
}

@article{Sun2021,
  doi = {10.1109/jstars.2021.3052869},
  url = {https://doi.org/10.1109/jstars.2021.3052869},
  year = {2021},
  publisher = {Institute of Electrical and Electronics Engineers ({IEEE})},
  volume = {14},
  pages = {2387--2402},
  author = {Xian Sun and Bing Wang and Zhirui Wang and Hao Li and Hengchao Li and Kun Fu},
  title = {Research Progress on Few-Shot Learning for Remote Sensing Image Interpretation},
  journal = {{IEEE} Journal of Selected Topics in Applied Earth Observations and Remote Sensing}
}

@article{Tong2020,
  doi = {10.1016/j.rse.2019.111322},
  url = {https://doi.org/10.1016/j.rse.2019.111322},
  year = {2020},
  month = feb,
  publisher = {Elsevier {BV}},
  volume = {237},
  pages = {111322},
  author = {Xin-Yi Tong and Gui-Song Xia and Qikai Lu and Huanfeng Shen and Shengyang Li and Shucheng You and Liangpei Zhang},
  title = {Land-cover classification with high-resolution remote sensing images using transferable deep models},
  journal = {Remote Sensing of Environment}
}

@article{Toth2016,
  doi = {10.1016/j.isprsjprs.2015.10.004},
  url = {https://doi.org/10.1016/j.isprsjprs.2015.10.004},
  year = {2016},
  month = may,
  publisher = {Elsevier {BV}},
  volume = {115},
  pages = {22--36},
  author = {Charles Toth and Grzegorz J{\'{o}}{\'{z}}k{\'{o}}w},
  title = {Remote sensing platforms and sensors: A survey},
  journal = {{ISPRS} Journal of Photogrammetry and Remote Sensing}
}

@article{Wang2020,
  doi = {10.1080/07038992.2020.1805729},
  url = {https://doi.org/10.1080/07038992.2020.1805729},
  year = {2020},
  month = aug,
  publisher = {Informa {UK} Limited},
  volume = {46},
  number = {5},
  pages = {501--531},
  author = {Yongzhi Wang and Hua Lv and Rui Deng and Shengbing Zhuang},
  title = {A Comprehensive Survey of Optical Remote Sensing Image Segmentation Methods},
  journal = {Canadian Journal of Remote Sensing}
}

@article{Wang2020b,
  doi = {10.3390/rs12020207},
  url = {https://doi.org/10.3390/rs12020207},
  year = {2020},
  month = jan,
  publisher = {{MDPI} {AG}},
  volume = {12},
  number = {2},
  pages = {207},
  author = {Sherrie Wang and William Chen and Sang Michael Xie and George Azzari and David B. Lobell},
  title = {Weakly Supervised Deep Learning for Segmentation of Remote Sensing Imagery},
  journal = {Remote Sensing}
}

@misc{wang2022loveda,
      title={LoveDA: A Remote Sensing Land-Cover Dataset for Domain Adaptive Semantic Segmentation}, 
      author={Junjue Wang and Zhuo Zheng and Ailong Ma and Xiaoyan Lu and Yanfei Zhong},
      year={2022},
      eprint={2110.08733},
      archivePrefix={arXiv},
      primaryClass={cs.CV}
}

@article{Wu2021,
  doi = {10.3390/rs13132582},
  url = {https://doi.org/10.3390/rs13132582},
  year = {2021},
  month = jul,
  publisher = {{MDPI} {AG}},
  volume = {13},
  number = {13},
  pages = {2582},
  author = {Zitong Wu and Biao Hou and Bo Ren and Zhongle Ren and Shuang Wang and Licheng Jiao},
  title = {A Deep Detection Network Based on Interaction of Instance Segmentation and Object Detection for {SAR} Images},
  journal = {Remote Sensing}
}

@misc{wu2023visual,
      title={Visual ChatGPT: Talking, Drawing and Editing with Visual Foundation Models}, 
      author={Chenfei Wu and Shengming Yin and Weizhen Qi and Xiaodong Wang and Zecheng Tang and Nan Duan},
      year={2023},
      eprint={2303.04671},
      archivePrefix={arXiv},
      primaryClass={cs.CV}
}

@inproceedings{Yang2015,
  doi = {10.1109/icdsp.2015.7251858},
  url = {https://doi.org/10.1109/icdsp.2015.7251858},
  year = {2015},
  month = jul,
  publisher = {{IEEE}},
  author = {Daiqin Yang and Zimeng Li and Yatong Xia and Zhenzhong Chen},
  title = {Remote sensing image super-resolution: Challenges and approaches},
  booktitle = {2015 {IEEE} International Conference on Digital Signal Processing ({DSP})}
}

@article{Yuan2020,
  doi = {10.1016/j.rse.2020.111716},
  url = {https://doi.org/10.1016/j.rse.2020.111716},
  year = {2020},
  month = may,
  publisher = {Elsevier {BV}},
  volume = {241},
  pages = {111716},
  author = {Qiangqiang Yuan and Huanfeng Shen and Tongwen Li and Zhiwei Li and Shuwen Li and Yun Jiang and Hongzhang Xu and Weiwei Tan and Qianqian Yang and Jiwen Wang and Jianhao Gao and Liangpei Zhang},
  title = {Deep learning in environmental remote sensing: Achievements and challenges},
  journal = {Remote Sensing of Environment}
}

@article{Yuan2021,
  doi = {10.1016/j.eswa.2020.114417},
  url = {https://doi.org/10.1016/j.eswa.2020.114417},
  year = {2021},
  month = may,
  publisher = {Elsevier {BV}},
  volume = {169},
  pages = {114417},
  author = {Xiaohui Yuan and Jianfang Shi and Lichuan Gu},
  title = {A review of deep learning methods for semantic segmentation of remote sensing imagery},
  journal = {Expert Systems with Applications}
}

@article{Zhang2020b,
  doi = {10.3390/rs12020221},
  url = {https://doi.org/10.3390/rs12020221},
  year = {2020},
  month = jan,
  publisher = {{MDPI} {AG}},
  volume = {12},
  number = {2},
  pages = {221},
  author = {Xiuwei Zhang and Jiaojiao Jin and Zeze Lan and Chunjiang Li and Minhao Fan and Yafei Wang and Xin Yu and Yanning Zhang},
  title = {{ICENET}: A Semantic Segmentation Deep Network for River Ice by Fusing Positional and Channel-Wise Attentive Features},
  journal = {Remote Sensing}
}

@misc{zhang2022dino,
      title={DINO: DETR with Improved DeNoising Anchor Boxes for End-to-End Object Detection}, 
      author={Hao Zhang and Feng Li and Shilong Liu and Lei Zhang and Hang Su and Jun Zhu and Lionel M. Ni and Heung-Yeung Shum},
      year={2022},
      eprint={2203.03605},
      archivePrefix={arXiv},
      primaryClass={cs.CV}
}

@misc{zhang2023personalize,
      title={Personalize Segment Anything Model with One Shot}, 
      author={Renrui Zhang and Zhengkai Jiang and Ziyu Guo and Shilin Yan and Junting Pan and Hao Dong and Peng Gao and Hongsheng Li},
      year={2023},
      eprint={2305.03048},
      archivePrefix={arXiv},
      primaryClass={cs.CV}
}

@misc{zhang2023visionlanguage,
      title={Vision-Language Models for Vision Tasks: A Survey}, 
      author={Jingyi Zhang and Jiaxing Huang and Sheng Jin and Shijian Lu},
      year={2023},
      eprint={2304.00685},
      archivePrefix={arXiv},
      primaryClass={cs.CV}
}

@misc{Zheng_2020_CVPR,
      title={Foreground-Aware Relation Network for Geospatial Object Segmentation in High Spatial Resolution Remote Sensing Imagery}, 
      author={Zhuo Zheng and Yanfei Zhong and Junjue Wang and Ailong Ma},
      year={2020},
      eprint={2011.09766},
      archivePrefix={arXiv},
      primaryClass={cs.CV}
}

@misc{zhang2022glipv2,
      title={GLIPv2: Unifying Localization and Vision-Language Understanding}, 
      author={Haotian Zhang and Pengchuan Zhang and Xiaowei Hu and Yen-Chun Chen and Liunian Harold Li and Xiyang Dai and Lijuan Wang and Lu Yuan and Jenq-Neng Hwang and Jianfeng Gao},
      year={2022},
      eprint={2206.05836},
      archivePrefix={arXiv},
      primaryClass={cs.CV}
}


\begin{thebibliography}{204}
\providecommand{\natexlab}[1]{#1}
\providecommand{\url}[1]{\texttt{#1}}
\expandafter\ifx\csname urlstyle\endcsname\relax
  \providecommand{\doi}[1]{doi: #1}\else
  \providecommand{\doi}{doi: \begingroup \urlstyle{rm}\Url}\fi

\bibitem{Adam2023}Adam, J., Liu, W., Zang, Y., Afzal, M., Bello, S., Muhammad, A., Wang, C. \& Li, J. Deep learning-based semantic segmentation of urban-scale 3D meshes in remote sensing: A survey. {\em International Journal Of Applied Earth Observation And Geoinformation}. \textbf{121} pp. 103365 (2023,7), https://doi.org/10.1016/j.jag.2023.103365
\bibitem{alayrac2022flamingo}Alayrac, J., Donahue, J., Luc, P., Miech, A., Barr, I., Hasson, Y., Lenc, K., Mensch, A., Millican, K., Reynolds, M., Ring, R., Rutherford, E., Cabi, S., Han, T., Gong, Z., Samangooei, S., Monteiro, M., Menick, J., Borgeaud, S., Brock, A., Nematzadeh, A., Sharifzadeh, S., Binkowski, M., Barreira, R., Vinyals, O., Zisserman, A. \& Simonyan, K. Flamingo: a Visual Language Model for Few-Shot Learning.  (2022)
\bibitem{Aleissaee2023}Aleissaee, A., Kumar, A., Anwer, R., Khan, S., Cholakkal, H., Xia, G. \& Khan, F. Transformers in Remote Sensing: A Survey. {\em Remote Sensing}. \textbf{15}, 1860 (2023,3), https://doi.org/10.3390/rs15071860
\bibitem{Amani2020}Amani, M., Ghorbanian, A., Ahmadi, S., Kakooei, M., Moghimi, A., Mirmazloumi, S., Moghaddam, S., Mahdavi, S., Ghahremanloo, M., Parsian, S., Wu, Q. \& Brisco, B. Google Earth Engine Cloud Computing Platform for Remote Sensing Big Data Applications: A Comprehensive Review. {\em IEEE Journal Of Selected Topics In Applied Earth Observations And Remote Sensing}. \textbf{13} pp. 5326-5350 (2020), https://doi.org/10.1109/jstars.2020.3021052
\bibitem{Bai2022}Bai, Y., Zhao, Y., Shao, Y., Zhang, X. \& Yuan, X. Deep learning in different remote sensing image categories and applications: status and prospects. {\em International Journal Of Remote Sensing}. \textbf{43}, 1800-1847 (2022,3), https://doi.org/10.1080/01431161.2022.2048319
\bibitem{BenjdiraRS2019}Benjdira, B., Bazi, Y., Koubaa, A. \& Ouni, K. Unsupervised Domain Adaptation Using Generative Adversarial Networks for Semantic Segmentation of Aerial Images. {\em Remote Sensing}. \textbf{11} (2019), https://doi.org/10.3390/rs11111369
\bibitem{Bressan2022}Bressan, P., Junior, J., Martins, J., Melo, M., Gonçalves, D., Freitas, D., Ramos, A., Furuya, M., Osco, L., Andrade Silva, J., Luo, Z., Garcia, R., Ma, L., Li, J. \& Gonçalves, W. Semantic segmentation with labeling uncertainty and class imbalance applied to vegetation mapping. {\em International Journal Of Applied Earth Observation And Geoinformation}. \textbf{108} pp. 102690 (2022,4)
\bibitem{boguszewski2022landcoverai}Boguszewski, A., Batorski, D., Ziemba-Jankowska, N., Dziedzic, T. \& Zambrzycka, A. LandCover.ai: Dataset for Automatic Mapping of Buildings, Woodlands, Water and Roads from Aerial Imagery.  (2022)
\bibitem{Chi2016}Chi, M., Plaza, A., Benediktsson, J., Sun, Z., Shen, J. \& Zhu, Y. Big Data for Remote Sensing: Challenges and Opportunities. {\em Proceedings Of The IEEE}. \textbf{104}, 2207-2219 (2016,11), https://doi.org/10.1109/jproc.2016.2598228
\bibitem{deCarvalho2022}Carvalho, O., Carvalho Júnior, O., Silva, C., Albuquerque, A., Santana, N., Borges, D., Gomes, R. \& Guimarães, R. Panoptic Segmentation Meets Remote Sensing. {\em Remote Sensing}. \textbf{14}, 965 (2022,2), https://doi.org/10.3390/rs14040965
\bibitem{dingAIDA2021}Ding, J., Xue, N., Xia, G., Bai, X., Yang, W., Yang, M., Belongie, S., Luo, J., Datcu, M., Pelillo, M. \& Zhang, L. Object Detection in Aerial Images: A Large-Scale Benchmark and Challenges. {\em IEEE Transactions On Pattern Analysis And Machine Intelligence}. pp. 1-1 (2021)
\bibitem{esa2023}European Space Agency SkySat - EOGateway.  (2023), https://earth.esa.int/eogateway/missions/SkySat, Accessed: 2023-05-29
\bibitem{Gao2021}Gao, K., Chen, M., Narges Fatholahi, S., He, H., Xu, H., Marcato Junior, J., Nunes Gonçalves, W., Chapman, M. \& Li, J. A region-based deep learning approach to instance segmentation of aerial orthoimagery for building rooftop extraction. {\em Geomatica}. \textbf{75}, 148-164 (2021)
\bibitem{Gharibbafghi2018}Gharibbafghi, Z., Tian, J. \& Reinartz, P. Modified Superpixel Segmentation for Digital Surface Model Refinement and Building Extraction from Satellite Stereo Imagery. {\em Remote Sensing}. \textbf{10}, 1824 (2018,11), https://doi.org/10.3390/rs10111824
\bibitem{Gmez2016}Gómez, C., White, J. \& Wulder, M. Optical remotely sensed time series data for land cover classification: A review. {\em ISPRS Journal Of Photogrammetry And Remote Sensing}. \textbf{116} pp. 55-72 (2016,6), https://doi.org/10.1016/j.isprsjprs.2016.03.008
\bibitem{Gonalves2023}Gonçalves, D., Marcato, J., Carrilho, A., Acosta, P., Ramos, A., Gomes, F., Osco, L., Rosa Oliveira, M., Martins, J., Damasceno, G., Araújo, M., Li, J., Roque, F., Faria Peres, L., Gonçalves, W. \& Libonati, R. Transformers for mapping burned areas in Brazilian Pantanal and Amazon with PlanetScope imagery. {\em International Journal Of Applied Earth Observation And Geoinformation}. \textbf{116} pp. 103151 (2023,2), https://doi.org/10.1016/j.jag.2022.103151
\bibitem{groundedsam2023}Grounded-Segment-Anything
. IDEA-Research. {\em GitHub Repository}. (2023), https://github.com/IDEA-Research/Grounded-Segment-Anything
\bibitem{Hossain2019}Hossain, M. \& Chen, D. Segmentation for Object-Based Image Analysis (OBIA): A review of algorithms and challenges from remote sensing perspective. {\em ISPRS Journal Of Photogrammetry And Remote Sensing}. \textbf{150} pp. 115-134 (2019,4), https://doi.org/10.1016/j.isprsjprs.2019.02.009
\bibitem{Hua2021}Hua, X., Wang, X., Rui, T., Shao, F. \& Wang, D. Cascaded panoptic segmentation method for high resolution remote sensing image. {\em Applied Soft Computing}. \textbf{109} pp. 107515 (2021,9), https://doi.org/10.1016/j.asoc.2021.107515
\bibitem{hua2021sparse}Hua, Y., Marcos, D., Mou, L., Zhu, X. \& Tuia, D. Semantic Segmentation of Remote Sensing Images with Sparse Annotations. {\em IEEE Geoscience And Remote Sensing Letters}. (2022)
\bibitem{Jozdani2022}Jozdani, S., Chen, D., Pouliot, D. \& Johnson, B. A review and meta-analysis of Generative Adversarial Networks and their applications in remote sensing. {\em International Journal Of Applied Earth Observation And Geoinformation}. \textbf{108} pp. 102734 (2022,4), https://doi.org/10.1016/j.jag.2022.102734
\bibitem{kirillov2023segment}Kirillov, A., Mintun, E., Ravi, N., Mao, H., Rolland, C., Gustafson, L., Xiao, T., Whitehead, S., Berg, A., Lo, W., Dollár, P. \& Girshick, R. Segment Anything.  (2023)
\bibitem{Kotaridis2021}Kotaridis, I. \& Lazaridou, M. Remote sensing image segmentation advances: A meta-analysis. {\em ISPRS Journal Of Photogrammetry And Remote Sensing}. \textbf{173} pp. 309-322 (2021,3), https://doi.org/10.1016/j.isprsjprs.2021.01.020
\bibitem{Li2020}Li, K., Hu, X., Jiang, H., Shu, Z. \& Zhang, M. Attention-Guided Multi-Scale Segmentation Neural Network for Interactive Extraction of Region Objects from High-Resolution Satellite Imagery. {\em Remote Sensing}. \textbf{12}, 789 (2020,3), https://doi.org/10.3390/rs12050789
\bibitem{li2022grounded}Li, L., Zhang, P., Zhang, H., Yang, J., Li, C., Zhong, Y., Wang, L., Yuan, L., Zhang, L., Hwang, J., Chang, K. \& Gao, J. Grounded Language-Image Pre-training. {\em CVPR}. (2022)
\bibitem{li2022uniformer}Li, K., Wang, Y., Zhang, J., Gao, P., Song, G., Liu, Y., Li, H. \& Qiao, Y. UniFormer: Unifying Convolution and Self-attention for Visual Recognition.  (2022)
\bibitem{Li2022}Li, X., Deng, J. \& Fang, Y. Few-Shot Object Detection on Remote Sensing Images. {\em IEEE Transactions On Geoscience And Remote Sensing}. \textbf{60} pp. 1-14 (2022), https://doi.org/10.1109/tgrs.2021.3051383
\bibitem{li2023uniformer}Li, K., Wang, Y., Zhang, J., Gao, P., Song, G., Liu, Y., Li, H. \& Qiao, Y. UniFormer: Unifying Convolution and Self-attention for Visual Recognition.  (2023)
\bibitem{li2023transformerbased}Li, X., Ding, H., Zhang, W., Yuan, H., Pang, J., Cheng, G., Chen, K., Liu, Z. \& Loy, C. Transformer-Based Visual Segmentation: A Survey.  (2023)
\bibitem{liu2023grounding}Liu, S., Zeng, Z., Ren, T., Li, F., Zhang, H., Yang, J., Li, C., Yang, J., Su, H., Zhu, J. \& Zhang, L. Grounding DINO: Marrying DINO with Grounded Pre-Training for Open-Set Object Detection.  (2023)
\bibitem{liu2023remoteclip}Liu, F., Chen, D., Guan, Z., Zhou, X., Zhu, J. \& Zhou, J. RemoteCLIP: A Vision Language Foundation Model for Remote Sensing.  (2023)
\bibitem{LoboTorres2020}Torres, D., Feitosa, R., Happ, P., Rosa, L., Junior, J., Martins, J., Bressan, P., Gonçalves, W. \& Liesenberg, V. Applying Fully Convolutional Architectures for Semantic Segmentation of a Single Tree Species in Urban Environment on High Resolution UAV Optical Imagery. {\em Sensors}. \textbf{20}, 563 (2020,1), https://doi.org/10.3390/s20020563
\bibitem{Lobry2020}Lobry, S., Marcos, D., Murray, J. \& Tuia, D. RSVQA: Visual Question Answering for Remote Sensing Data. {\em IEEE Transactions On Geoscience And Remote Sensing}. \textbf{58}, 8555-8566 (2020,12), https://doi.org/10.1109/tgrs.2020.2988782
\bibitem{loshchilov2017sgdr}Loshchilov, I. \& Hutter, F. SGDR: Stochastic Gradient Descent with Warm Restarts.  (2017)
\bibitem{Lu2021}Lu, J., Jia, H., Li, T., Li, Z., Ma, J. \& Zhu, R. An Instance Segmentation Based Framework for Large-Sized High-Resolution Remote Sensing Images Registration. {\em Remote Sensing}. \textbf{13}, 1657 (2021,4), https://doi.org/10.3390/rs13091657
\bibitem{Ma2022}Ma, A., Wang, J., Zhong, Y. \& Zheng, Z. FactSeg: Foreground Activation-Driven Small Object Semantic Segmentation in Large-Scale Remote Sensing Imagery. {\em IEEE Transactions On Geoscience And Remote Sensing}. \textbf{60} pp. 1-16 (2022), https://doi.org/10.1109/tgrs.2021.3097148
\bibitem{mai2023opportunities}Mai, G., Huang, W., Sun, J., Song, S., Mishra, D., Liu, N., Gao, S., Liu, T., Cong, G., Hu, Y., Cundy, C., Li, Z., Zhu, R. \& Lao, N. On the Opportunities and Challenges of Foundation Models for Geospatial Artificial Intelligence.  (2023)
\bibitem{Martins2021}Martins, J., Nogueira, K., Osco, L., Gomes, F., Furuya, D., Gonçalves, W., Sant'Ana, D., Ramos, A., Liesenberg, V., Santos, J., Oliveira, P. \& Junior, J. Semantic Segmentation of Tree-Canopy in Urban Environment with Pixel-Wise Deep Learning. {\em Remote Sensing}. \textbf{13}, 3054 (2021,8), https://doi.org/10.3390/rs13163054
\bibitem{mialon2023augmented}Mialon, G., Dessì, R., Lomeli, M., Nalmpantis, C., Pasunuru, R., Raileanu, R., Rozière, B., Schick, T., Dwivedi-Yu, J., Celikyilmaz, A., Grave, E., LeCun, Y. \& Scialom, T. Augmented Language Models: a Survey.  (2023)
\bibitem{Minaee2021}Minaee, S., Boykov, Y., Porikli, F., Plaza, A., Kehtarnavaz, N. \& Terzopoulos, D. Image Segmentation Using Deep Learning: A Survey. {\em IEEE Transactions On Pattern Analysis And Machine Intelligence}. pp. 1-1 (2021), https://doi.org/10.1109/tpami.2021.3059968
\bibitem{openai2023gpt4}OpenAI. GPT-4 Technical Report. (2023)
\bibitem{Osco2020}Osco, L., Arruda, M., Junior, J., Silva, N., Ramos, A., Akemi Saito Moryia, Imai, N., Pereira, D., Creste, J., Matsubara, E., Li, J. \& Gonçalves, W. A convolutional neural network approach for counting and geolocating citrus-trees in UAV multispectral imagery. {\em ISPRS Journal Of Photogrammetry And Remote Sensing}. \textbf{160} pp. 97-106 (2020,2), https://doi.org/10.1016/j.isprsjprs.2019.12.010
\bibitem{Osco2021_cnn}Osco, L., Arruda, M., Gonçalves, D., Dias, A., Batistoti, J., Souza, M., Gomes, F., Ramos, A., Castro Jorge, L., Liesenberg, V., Li, J., Ma, L., Marcato, J. \& Gonçalves, W. A CNN approach to simultaneously count plants and detect plantation-rows from UAV imagery. {\em ISPRS Journal Of Photogrammetry And Remote Sensing}. \textbf{174} pp. 1-17 (2021,4), https://doi.org/10.1016/j.isprsjprs.2021.01.024
\bibitem{Osco2021}Osco, L., Junior, J., Ramos, A., Castro Jorge, L., Fatholahi, S., Andrade Silva, J., Matsubara, E., Pistori, H., Gonçalves, W. \& Li, J. A review on deep learning in UAV remote sensing. {\em International Journal Of Applied Earth Observation And Geoinformation}. \textbf{102} pp. 102456 (2021,10), https://doi.org/10.1016/j.jag.2021.102456
\bibitem{Osco2021segmentation}Osco, L., Nogueira, K., Ramos, A., Pinheiro, M., Furuya, D., Gonçalves, W., Castro Jorge, L., Junior, J. \& Santos, J. Semantic segmentation of citrus-orchard using deep neural networks and multispectral UAV-based imagery. {\em Precision Agriculture}. \textbf{22}, 1171-1188 (2021,1), https://doi.org/10.1007/s11119-020-09777-5
\bibitem{Osco2023}Osco, L., Lemos, E., Gonçalves, W., Ramos, A. \& Junior, J. The Potential of Visual ChatGPT for Remote Sensing. {\em Remote Sensing}. \textbf{15}, 3232 (2023,6), https://doi.org/10.3390/rs15133232
\bibitem{AIRemoteSensing2023}Osco, L. AI-RemoteSensing: a collection of Jupyter and Google Colaboratory notebooks dedicated to leveraging Artificial Intelligence (AI) in Remote Sensing applications. (Zenodo,2023,6), https://doi.org/10.5281/zenodo.8092269
\bibitem{powers2020evaluation}Powers, D. Evaluation: from precision, recall and F-measure to ROC, informedness, markedness and correlation.  (2020)
\bibitem{Qurratulain2023}Qurratulain, S., Zheng, Z., Xia, J., Ma, Y. \& Zhou, F. Deep learning instance segmentation framework for burnt area instances characterization. {\em International Journal Of Applied Earth Observation And Geoinformation}. \textbf{116} pp. 103146 (2023,2)
\bibitem{Rahnemoonfar2021}Rahnemoonfar, M., Chowdhury, T., Sarkar, A., Varshney, D., Yari, M. \& Murphy, R. FloodNet: A High Resolution Aerial Imagery Dataset for Post Flood Scene Understanding. {\em IEEE Access}. \textbf{9} pp. 89644-89654 (2021), https://doi.org/10.1109/access.2021.3090981
\bibitem{Rahman2016}Rahman, M. \& Wang, Y. Optimizing Intersection-Over-Union in Deep Neural Networks for Image Segmentation. {\em Advances In Visual Computing}. pp. 234-244 (2016), https://doi.org/10.1007/978-3-319-50835-1\_22
\bibitem{Song2023}Song, Y., Kalacska, M., Gašparović, M., Yao, J. \& Najibi, N. Advances in geocomputation and geospatial artificial intelligence (GeoAI) for mapping. {\em International Journal Of Applied Earth Observation And Geoinformation}. \textbf{120} pp. 103300 (2023,6)
\bibitem{Su2019}Su, H., Wei, S., Yan, M., Wang, C., Shi, J. \& Zhang, X. Object Detection and Instance Segmentation in Remote Sensing Imagery Based on Precise Mask R-CNN. {\em IGARSS 2019 - 2019 IEEE International Geoscience And Remote Sensing Symposium}. (2019,7), https://doi.org/10.1109/igarss.2019.8898573
\bibitem{Sun2021}Sun, X., Wang, B., Wang, Z., Li, H., Li, H. \& Fu, K. Research Progress on Few-Shot Learning for Remote Sensing Image Interpretation. {\em IEEE Journal Of Selected Topics In Applied Earth Observations And Remote Sensing}. \textbf{14} pp. 2387-2402 (2021), https://doi.org/10.1109/jstars.2021.3052869
\bibitem{Tong2020}Tong, X., Xia, G., Lu, Q., Shen, H., Li, S., You, S. \& Zhang, L. Land-cover classification with high-resolution remote sensing images using transferable deep models. {\em Remote Sensing Of Environment}. \textbf{237} pp. 111322 (2020,2), https://doi.org/10.1016/j.rse.2019.111322
\bibitem{Toth2016}Toth, C. \& Jóźków, G. Remote sensing platforms and sensors: A survey. {\em ISPRS Journal Of Photogrammetry And Remote Sensing}. \textbf{115} pp. 22-36 (2016,5), https://doi.org/10.1016/j.isprsjprs.2015.10.004
\bibitem{Wang2020}Wang, Y., Lv, H., Deng, R. \& Zhuang, S. A Comprehensive Survey of Optical Remote Sensing Image Segmentation Methods. {\em Canadian Journal Of Remote Sensing}. \textbf{46}, 501-531 (2020,8), https://doi.org/10.1080/07038992.2020.1805729
\bibitem{Wang2020b}Wang, S., Chen, W., Xie, S., Azzari, G. \& Lobell, D. Weakly Supervised Deep Learning for Segmentation of Remote Sensing Imagery. {\em Remote Sensing}. \textbf{12}, 207 (2020,1), https://doi.org/10.3390/rs12020207
\bibitem{wang2022loveda}Wang, J., Zheng, Z., Ma, A., Lu, X. \& Zhong, Y. LoveDA: A Remote Sensing Land-Cover Dataset for Domain Adaptive Semantic Segmentation.  (2022)
\bibitem{segment-geospatial2023}Wu, Q., Osco, L. samgeo: A Python package for segmenting geospatial data with the Segment Anything Model (SAM). (Zenodo,2023,5), https://doi.org/10.5281/zenodo.7966658
\bibitem{Wu2021}Wu, Z., Hou, B., Ren, B., Ren, Z., Wang, S. \& Jiao, L. A Deep Detection Network Based on Interaction of Instance Segmentation and Object Detection for SAR Images. {\em Remote Sensing}. \textbf{13}, 2582 (2021,7), https://doi.org/10.3390/rs13132582
\bibitem{wu2023visual}Wu, C., Yin, S., Qi, W., Wang, X., Tang, Z. \& Duan, N. Visual ChatGPT: Talking, Drawing and Editing with Visual Foundation Models.  (2023)
\bibitem{Xu2023}Xu, Y., Bai, T., Yu, W., Chang, S., Atkinson, P. \& Ghamisi, P. AI Security for Geoscience and Remote Sensing: Challenges and future trends. {\em IEEE Geoscience And Remote Sensing Magazine}. \textbf{11}, 60-85 (2023,6), https://doi.org/10.1109/mgrs.2023.3272825
\bibitem{Yang2015}Yang, D., Li, Z., Xia, Y. \& Chen, Z. Remote sensing image super-resolution: Challenges and approaches. {\em 2015 IEEE International Conference On Digital Signal Processing (DSP)}. (2015,7), https://doi.org/10.1109/icdsp.2015.7251858
\bibitem{Yuan2020}Yuan, Q., Shen, H., Li, T., Li, Z., Li, S., Jiang, Y., Xu, H., Tan, W., Yang, Q., Wang, J., Gao, J. \& Zhang, L. Deep learning in environmental remote sensing: Achievements and challenges. {\em Remote Sensing Of Environment}. \textbf{241} pp. 111716 (2020,5), https://doi.org/10.1016/j.rse.2020.111716
\bibitem{Yuan2021}Yuan, X., Shi, J. \& Gu, L. A review of deep learning methods for semantic segmentation of remote sensing imagery. {\em Expert Systems With Applications}. \textbf{169} pp. 114417 (2021,5), https://doi.org/10.1016/j.eswa.2020.114417
\bibitem{Zhang2020b}Zhang, X., Jin, J., Lan, Z., Li, C., Fan, M., Wang, Y., Yu, X. \& Zhang, Y. ICENET: A semantic segmentation deep network for river ice by fusing positional and channel-wise attentive features. {\em Remote Sensing}. \textbf{12}, 1-22 (2020)
\bibitem{Zhang2021}Zhang, R., Li, G., Wunderlich, T. \& Wang, L. A survey on deep learning-based precise boundary recovery of semantic segmentation for images and point clouds. {\em International Journal Of Applied Earth Observation And Geoinformation}. \textbf{102} pp. 102411 (2021,10), https://doi.org/10.1016/j.jag.2021.102411
\bibitem{zhang2022dino}Zhang, H., Li, F., Liu, S., Zhang, L., Su, H., Zhu, J., Ni, L. \& Shum, H. DINO: DETR with Improved DeNoising Anchor Boxes for End-to-End Object Detection.  (2022)
\bibitem{zhang2023personalize}Zhang, R., Jiang, Z., Guo, Z., Yan, S., Pan, J., Dong, H., Gao, P. \& Li, H. Personalize Segment Anything Model with One Shot.  (2023)
\bibitem{zhang2023visionlanguage}Zhang, J., Huang, J., Jin, S. \& Lu, S. Vision-Language Models for Vision Tasks: A Survey.  (2023)
\bibitem{Zheng_2020_CVPR}Zheng, Z., Zhong, Y., Wang, J. \& Ma, A. Foreground-Aware Relation Network for Geospatial Object Segmentation in High Spatial Resolution Remote Sensing Imagery. {\em Proceedings Of The IEEE/CVF Conference On Computer Vision And Pattern Recognition (CVPR)}. (2020,6)
\bibitem{zhang2022glipv2}Zhang, H., Zhang, P., Hu, X., Chen, Y., Li, L., Dai, X., Wang, L., Yuan, L., Hwang, J. \& Gao, J. GLIPv2: Unifying Localization and Vision-Language Understanding. {\em ArXiv Preprint ArXiv:2206.05836}. (2022)
\bibitem{Zhang2023}Zhang, Z., Zhang, Q., Hu, X., Zhang, M. \& Zhu, D. On the automatic quality assessment of annotated sample data for object extraction from remote sensing imagery. {\em ISPRS Journal Of Photogrammetry And Remote Sensing}. \textbf{201} pp. 153-173 (2023,7), https://doi.org/10.1016/j.isprsjprs.2023.05.026
\bibitem{Zia2022}Zia, U., Riaz, M. \& Ghafoor, A. Transforming remote sensing images to textual descriptions. {\em International Journal Of Applied Earth Observation And Geoinformation}. \textbf{108} pp. 102741 (2022,4)

\end{thebibliography}


\end{document}